\documentclass[conference]{IEEEtran}
\IEEEoverridecommandlockouts
\usepackage{cite}
\usepackage{amsmath,amssymb,amsfonts}
\usepackage{algorithm,algorithmic}
\usepackage{graphicx}
\usepackage{textcomp}
\usepackage{subcaption}
\usepackage{cleveref}
\usepackage{xcolor}
\usepackage[hang, flushmargin]{footmisc} 
\usepackage{adjustbox}

\def\BibTeX{{\rm B\kern-.05em{\sc i\kern-.025em b}\kern-.08em
    T\kern-.1667em\lower.7ex\hbox{E}\kern-.125emX}}

\newcommand{\name}{ProspectNet}
    
\begin{document}

\title{\name{}: Weighted Conditional Attention for Future Interaction Modeling in Behavior Prediction\\

\thanks{
\textsuperscript{\dag}Work done during an internship at Xsense.ai Inc.\\
\textsuperscript{1}School for Engineering of Matter, Transport \& Energy, Arizona State University. 975 S. Myrtle Ave, Tempe, AZ 85281. \texttt{Yutian.Pang@asu.edu} \\
\textsuperscript{2}Behavior Prediction Team, Xsense.ai Inc. 15070 Avenue of Science \#100, San Diego, CA 92128. \texttt{\{zehuag1, binnanz\}@xiaopeng.com} }
}

\author{
Yutian Pang\textsuperscript{1\dag},
Zehua Guo\textsuperscript{2},
Binnan Zhuang\textsuperscript{2}\\
}

\maketitle

\begin{abstract}
Behavior prediction plays an important role in integrated autonomous driving software solutions. In behavior prediction research, interactive behavior prediction is a less-explored area, compared to single-agent behavior prediction. Predicting the motion of interactive agents requires initiating novel mechanisms to capture the joint behaviors of the interactive pairs. In this work, we formulate the end-to-end joint prediction problem as a sequential learning process of marginal learning and joint learning of vehicle behaviors. We propose \name{}, a joint learning block that adopts the weighted attention score to model the mutual influence between interactive agent pairs. The joint learning block first weighs the multi-modal predicted candidate trajectories, then updates the ego-agent's embedding via cross attention. Furthermore, we broadcast the individual future predictions for each interactive agent into a pair-wise scoring module to select the top $K$ prediction pairs. We show that \name{} outperforms the Cartesian product of two marginal predictions, and achieves comparable performance on the Waymo Interactive Motion Prediction benchmarks.
\end{abstract}


\section{Introduction\label{introduction}}
Along with perception and planning, behavior prediction forms an integrated end-to-end autonomous driving system (ADS). In ADS, behavior prediction infers the future locations of nearby dynamic objects identified by perception systems, and the maneuver of the ego vehicle is further scheduled by planning systems. The task of behavior prediction is challenging due to complex operation scenarios such as vehicle-to-vehicle, vehicle-to-human, and vehicle-to-road interactions under the various right-of-ways and driver intentions.

\begin{figure}[htbp]
\centering
\includegraphics[width=0.45\textwidth]{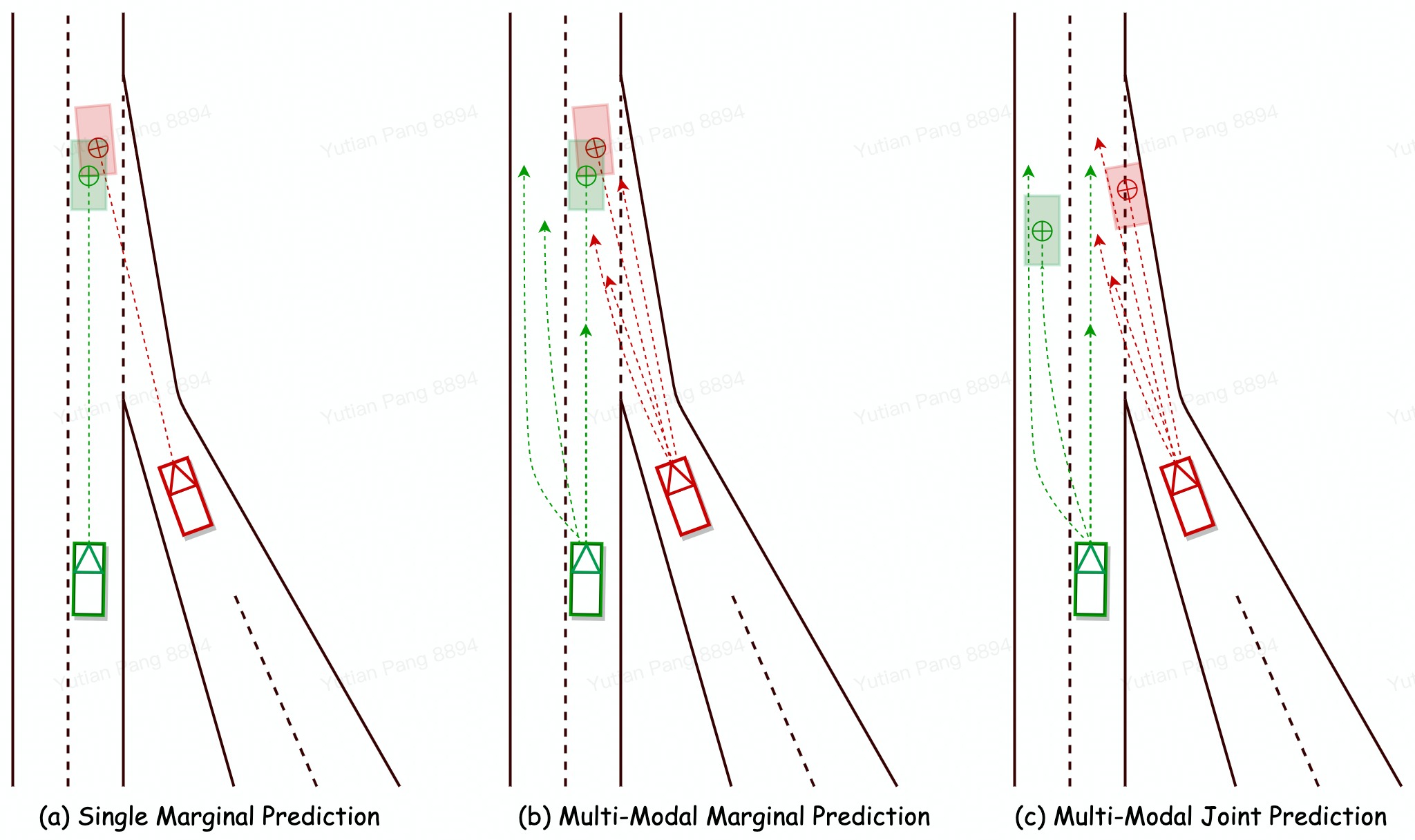}
\caption{A demonstration of different behavior prediction problems in a merging scenario. In single marginal prediction, the model predicts one future trajectory sequence for each agent without considering interactions. In multi-modal marginal prediction, the model gives multiple future predictions for each agent without considering the interactions. In multi-modal joint prediction, the model considers the collisions and selects the best potential routes. In multimodality settings, $\oplus$ marks the mode with the highest probability.}
\label{fig: bp-demo}
\end{figure}

In behavior prediction, researchers are focusing on several types of problems, as shown in Fig.~\ref{fig: bp-demo}. We name the model for predicting the trajectory of a single agent without considering interacting agents a single \textit{marginal model}. In motion prediction literature, multimodality has become the standard practice to capture uncertainties into the future \cite{sun2022m2i,ngiam2021scene, liu2021multimodal, casas2020importance}, where the marginal predictor outputs a set of trajectories (modes) to represent different driver behavior intentions. Each trajectory sequence is associated with a corresponding probability (mode probability) as the prediction confidence score. To overcome the issue of mode collapse, \textit{post-hoc} processing methods such as non-maximum suppression \cite{zhao2020tnt} filtering are proposed to reject near-duplicated trajectories. In this work, we are particularly interested in the highly-interactive scenario of the prediction systems of Autonomous Vehicles (AVs), and this leads to the joint prediction of interactive vehicles, where multiple realizations of self-consistent future trajectories are generated. In our multi-modal joint prediction, the predictions of one agent are \textit{conditionally inferred} \cite{tolstaya2021identifying} on other agents' predictions and vice versa. Therefore, our joint model learns to generate predictions in a sense that are aware of both past and future interactions. 

Many recent publications adopt the graph-based feature extraction for interaction representation \cite{mo2021multi,zhao2020tnt,gao2020vectornet}. VectorNet \cite{gao2020vectornet} encodes high-precision map polylines as graph features and represents interactions via lower-order self-interactions and higher-order agent graph interactions. Target-driven trajectory prediction network \cite{zhao2020tnt} integrates VectorNet's interaction representation idea and further extends to multi-modal sense by pre-sampling target points as anchors of predictions. HEAT adopts edge-enhanced graph attention to model the interaction \cite{mo2021multi}. M2I \cite{sun2022m2i} proposes a sequential procedure for interactive motion prediction, which includes a classification of influencer-reactor relations, and then a conditional predictor to generate predictions for the reactor. 

In this work, we propose an end-to-end trajectory prediction framework for interactive motion prediction, build upon the fine-tuned marginal motion prediction model. Similar to the marginal model, the proposed model takes embedded inputs from both dynamic and static--moving objects histories (cars, pedestrians, cyclists), roadmaps, etc. To bridge the gap between a marginal motion prediction and an interactive motion prediction, we propose a novel weighted attention mechanism for condition prediction of interactive agent trajectories and named it \name{}. One of our innovation is the introduction of a weight bias term to the compatibility function of simple attention mechanism. The weight bias term represents the conditional importance of predicted trajectory candidates from the interactive agent, so the joint encoding module can learn the probability distribution of the marginal prediction as a priori. A uniform probability importance score for each lane segment is also introduced for normalization. Another important building block of the end-to-end joint model is the pair-wise scoring module to \textit{post-hoc} rank the paired trajectory candidates. 

Our work explicitly models the interaction between interacting agents, by a specially designed attention mechanism, with a pair-wise trajectory candidates selection function for candidate selection. We examine the effectiveness of the proposed framework by comparing it with a fine-tuned marginal prediction model (Cartesian product between topK candidates of two agents). The usage of joint learning block is flexible and can be extended to stacked interactions and multi-agents. Different from existing literature, we do not have a heuristic-based pre-labeling procedure for \textit{influencer/reactor} with different types of relations. Moreover, we only use the vectorized features in our implementation. To be more clear, we highlight our contributions here: (a) We propose a joint learning block where the weighted conditional attention mechanism is the core component. The joint learning block is introduced to the prediction embedding of the marginal model, as a sequential process. 
(b) We propose a pair-wise scoring module to select broadcasted topK candidate pairs from top $N^2$ pairs. (c) We show the marginal model largely improves the prediction capability in various scenarios through visualizations.

\section{Literature Review\label{review}}
Trajectory prediction is a popular research topic in various engineering application domains including cars, pedestrians \cite{yu2020spatio}, ships \cite{perera2012maritime}, aircraft \cite{pang2022bayesian}, etc. In this section, we will discuss the extensive motion prediction research for ground-based autonomous systems. We will start from common practices in motion prediction \Cref{motion}, then extend to agent interaction modeling for conditional behavior prediction in \Cref{cbp}. 

\subsection{Motion Prediction in Behavior Prediction\label{motion}}
\textbf{Context Encoding} acts as the initial step to extract features from raw input data, with deep neural networks (e.g., CNN \cite{cui2019multimodal,casas2018intentnet}, RNNs \cite{mo2021multi}, GNNs \cite{gao2020vectornet,zeng2021lanercnn}) for unsupervised feature extraction. The context encoding methods can be divided into two categories, rasterized methods and vectorized methods. Rasterized encoding treats the HD map and agent history tracks as images and adopts a CNN-based encoder to construct inputs to trajectory predictor/decoder. A typical practice is to represent timestamps as image kernels. Multipath and its variant \cite{chai2019multipath,varadarajan2021multipath++} use CNNs-based feature encoder for raster images. IntentNet \cite{casas2018intentnet} develops CNN-based feature encoder for both raster images and LiDAR points. DRF-Net \cite{jain2020discrete} rasterizes map elements and augments agent encoding as inputs to the recursive discrete residual flow module. In contrast, sparse encoding (vectorized) methods treat each entity as a structured element set and adopt graph-based algorithm to extract features and learn element interactions. Vectorized methods develop rapidly with benefits from reduced training difficulty of fewer model parameters, and explicit scenario encoding for improved explainability in structured feature representations. VectorNet \cite{gao2020vectornet} is the pioneering research in this direction. It explicitly encodes the sparse element representations of agents/maps as graph features and adopts lower and higher-order interaction learning modules for modeling agent-to-agent and agent-to-map relations. TNT \cite{zhao2020tnt} and DenseTNT \cite{gu2021densetnt} further extend this concept by using sparse/dense anchor/goal-based trajectory predictors and considering multi-modality of driving intent. LaneGCN and LaneRCNN \cite{liang2020learning,zeng2021lanercnn} focus on modeling lane graph interactions and graph-to-graph interactions. Additionally, M2I \cite{sun2022m2i} proposes to combine rasterized encoding and vectorized encoding for better feature representation and achieve state-of-the-art.

\textbf{Target-Driven Trajectory Prediction} TNT introduces pre-sampled anchors from map segments and predicts trajectories based on anchors with probabilities. TNT demonstrates the efficiency of using the non-maximum suppression (NMS) \cite{zhao2020tnt} threshold to reject near-duplicate trajectories, such that diverse intents of the driver can be filtered from trajectory candidates. However, the sparse anchors in TNT fail to perfectly estimate the probability of roads. DenseTNT \cite{gu2021densetnt} improves the sparse goal-sampling procedure with dense goal probability estimation on the map. In DenseTNT, the dense goals are selected along the candidate road lanes that are classified based on the distance between goals and lanes. In the goal training stage, the goal probability is predicted through attention. The predicted trajectories will be based on the topK selected goal candidates. HOME \cite{gilles2021home} uses a similar goal-sampling idea but uses CNN to generate the probability heatmap. Generally, the goal-based idea has been explored in different settings toward various objectives \cite{zeng2021lanercnn,choi2019drogon}.

\textbf{Multi-Modality} is an approach to provoke uncertainty in behavior prediction. Multi-modal prediction aims to generate multiple plausible trajectories for the objective. Multi-modal prediction is designed to handle uncertainty in motion prediction. The uncertainty comes from diverse behaviors of drivers, and dynamic traffic scenarios. In practice, multi-modal prediction is achieved by either randomized latent variables \cite{tang2019multiple}, or directly sampled from probability distributions (e.g., GMM \cite{khandelwal2020if,tolstaya2021identifying}) to get diverse predictions. These methods are sensitive to the model's pre-defined parameters/distributions. Alternatively, anchor/goal-based approaches make predictions based on the proposals and reduce computational difficulties by narrowing down the solution search space. We follow the second approach in this work. 

\begin{figure*}[htbp]
\centering
\includegraphics[width=0.95\textwidth]{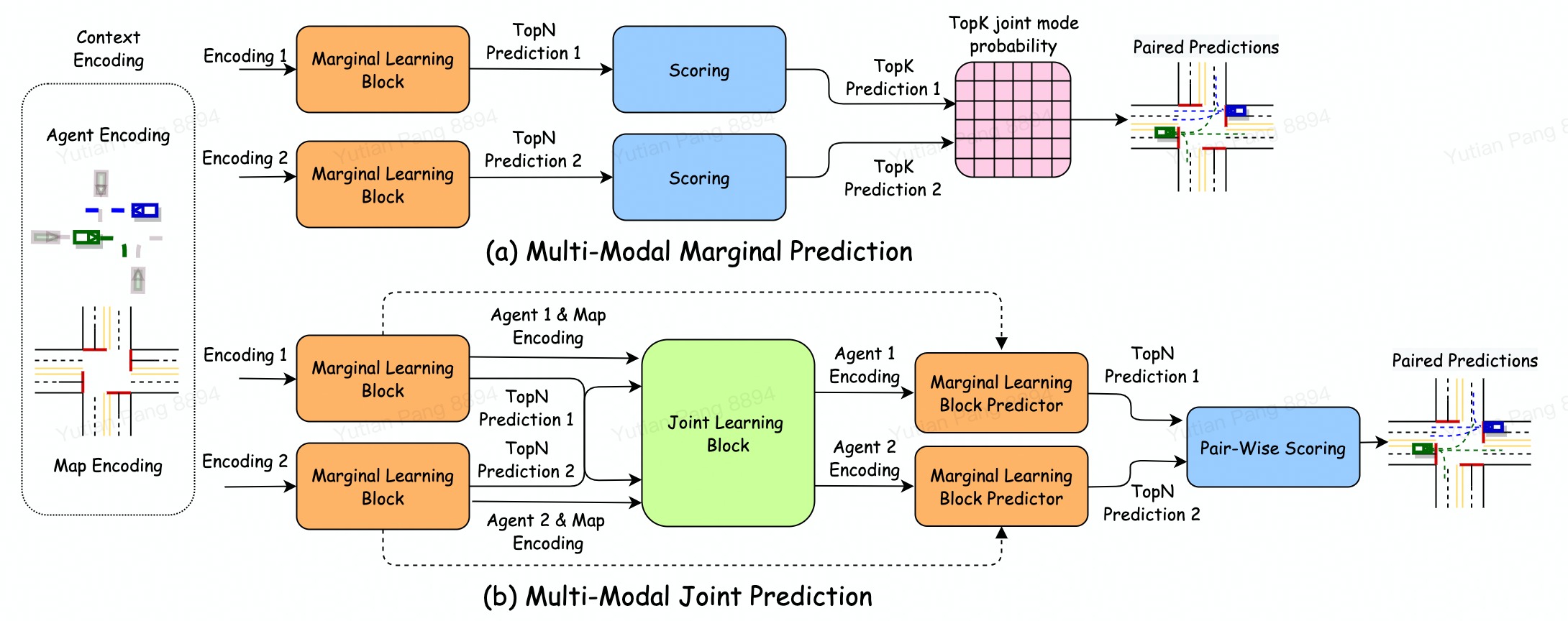}
\caption{\textbf{Overview of the \name{}}: In (a), the interactive pair predictions from the marginal prediction model. First, The marginal model parameter is fine-tuned. Then, the topK candidates from topN candidates of each agent will be scored. Finally, the topK pair candidate trajectories will be selected from the Cartesian product of $K^2$ paired candidates. In (b), the joint learning block updates each agent's marginal encoding with a weighted attention mechanism. The updated agent encoding is further used to sample N candidates of each agent. The pair-wise scoring block selects topK pair candidate trajectories.}
\label{fig: overview}
\end{figure*}

\subsection{Interaction Modeling in Conditional Behavior Prediction\label{cbp}}
Predicting trajectories considering interactions between two or more agents remains an open question in the topic of prediction. Existing methods can be divided into classical methods and deep learning based methods. Classical methods (e.g., Social Force models, Geometry-based methods) require hand-crafted features to capture multi-agent behaviors. They are less data-intensive with increased interpretability. Social Force models (guided by virtual repulsive and attractive forces) are built upon the assumption that pedestrians are mission-driven for destination navigation and collision avoidance \cite{helbing1995social}. However, Social Force models perform poorly on pedestrian prediction tasks \cite{kuderer2012feature}. Geometry-based models adopt optimization-based interactive prediction with the geometry of each agent \cite{luo2018porca, van2011reciprocal}. Classical methods require extensive feature engineering and are hard to generalize in different scenes \cite{huang2019stgat}. Deep learning based methods show satisfactory performance compared with classical methods. Behavior CNN captures crowd behaviors using CNN \cite{yi2016pedestrian}. Social pooling merges the latent space between nearby agents, which leads to a socially-aware prediction. GNNs, self-attention, and Transformer mechanisms are frequently introduced to leverage agent interactions \cite{tang2019multiple,kamra2020multi,ngiam2021scene,li2020end,kosaraju2019social}. Conditional predictions output future agent trajectories by conditioning on the nearby agent's planned or future trajectories. 

In this work, we propose \name{}, a novel attention mechanism to explicitly represent conditional inference, in multi-modality fashion. \name{} adopts attention to model interactions between interactive agent pairs. The prediction of one agent is conditioned on all trajectory candidates of another agent, with adjusted attention score bias to avoid restricted search space and to represent candidate confidence. Different from the conditional trajectory predictor in M2I, our proposed model doesn't have a pre-labeled influencer and reactor, yet the joint learning block is newly proposed sequentially aligned after the marginal trajectory predictor.

\section{Methodology\label{methods}}
In this section, we discuss the problem setup and an overview of \name{} for the interactive motion prediction case. As shown in Fig.~\ref{fig: overview}, our proposed framework contains a marginal learning block, a joint learning block, and a pair-wise scoring module. We adopt a similar structure to TNT \cite{zhao2020tnt} as the baseline marginal learning block. For the simplicity of illustration, we focus on the case of two interactive agents, numbered 1 and 2. Firstly, each agent's map and trajectory encoding will be passed into the marginal learning block separately. The marginal learning block will output each agent's top $N$ trajectory candidates, future trajectory embedding, and map embedding. Then, the joint learning block takes agent A's future trajectory embedding, agent B's prediction candidates, agent A's map embedding, and output updated agent A's future embedding. The embedding is further used to sample trajectory candidates from the predictor of marginal learning block. Lastly, the top $N$ trajectory candidates for each agent are combined into $N^2$ pairs to be selected by the pair-wise scoring module. The final output dimension will be $\mathbb{R}^{2\times K\times T\times 2}$, where K is the number of modes, T is prediction timestamps, and the last dimension is the coordinate dimension (2 in this work).

\subsection{Problem Formulation\label{problem}}
\begin{figure}[htbp]
    \centering
    \includegraphics[width=0.4\textwidth]{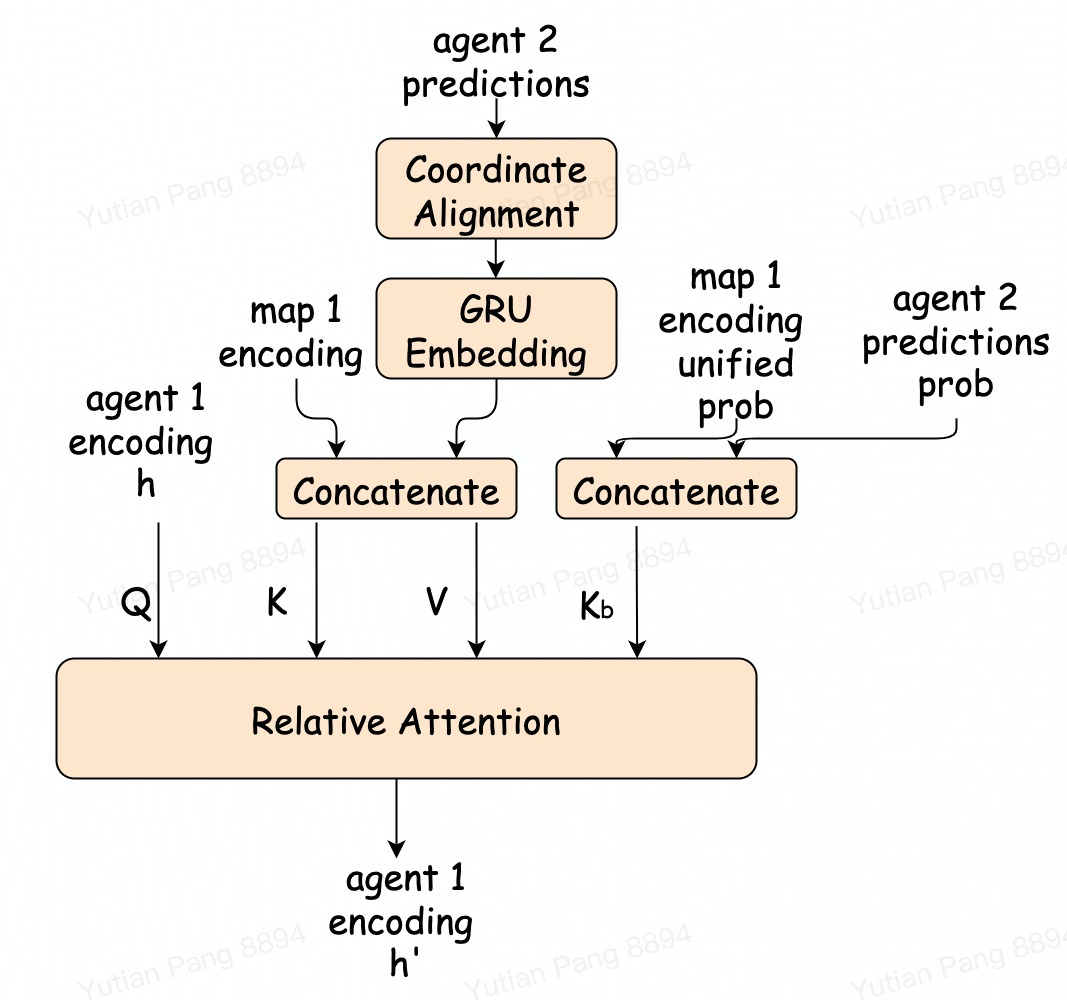}
    \caption{Joint Learning Block (demonstration with agent A only): We propose relative attention to consider the attention score. The query (Q) matrix is the agent encoding. The key (K) and value (V) metrics are agents A map encoding and aligned agent B topN predictions. The weight for attention score is the unified agent A map encoding probability and agent B topN candidates probability.}
    \label{fig: jointblock}
\end{figure}

The objective of interactive motion prediction is to predict two agents ($A\& B$) future $T$ timestamps trajectories collaboratively. Researchers have proposed several methods to determine if the two agents are considered interactive. For instance, CBP \cite{tolstaya2021identifying} proposes to use \textit{mutual information} to quantify the dependence between two time series. Several motion prediction datasets explicitly provide the interactive label for agents in a scene \cite{zhan2019interaction,ettinger2021large}. In this work, we primarily focus on the interactive bahavior prediction problem for the Waymo Motion Prediction Dataset. For a given scenario, the future trajectory sequences for agent A are $\mathbf{S_A} = \{\mathbf{s_A^i}\in \mathbb{R}^{T\times 2}: i\in \{1, 2,..., K\}\}$ for top $K$ candidates, or $i\in \{1, 2, ..., N\}$ for top $N$ candidates before candidate selection. We use $\mathbf{x_A}=\{\mathbf{m_A}, \mathbf{c_A}\}$ to represent context encoding, which consists of the map embedding $\mathbf{m_A}\in \mathbb{R}^{H\times E}$ and agent embedding $\mathbf{c_A}\in \mathbb{R}^{T\times E}$, where $H$ is the number of features, and $E$ is the embedding dimension. We use $\mathbf{p_A} \in \mathbb{R}^{K\times 1}$ to denote the associate probability to each candidate of $\mathbf{S_A}$.

\begin{algorithm}
    \caption{Joint Learning Process}
    \label{algo: jointlearning}
    \begin{algorithmic}[1]
    \renewcommand{\algorithmicrequire}{\textbf{Input:}}
    \renewcommand{\algorithmicensure}{\textbf{Objective:}}
    \REQUIRE Agent embedding $\mathbf{h_A}$, $\mathbf{h_B}$; Context encoding each coordinate $\mathbf{x_A}$, $\mathbf{x_B}$; Predicted trajectories $\mathbf{S_A}$, $\mathbf{S_B}$ and associated probabilities $\mathbf{p_A}$, $\mathbf{p_A}$.
    \ENSURE  Updated agent encoding $\mathbf{\hat{h}_A}$, $\mathbf{\hat{h}_B}$
        \FOR {$q$ in $0$ to $Q$}
        \STATE Coordinate alignment: $\mathbf{\hat{s_B}} \leftarrow \mathbf{s_B}$
        \STATE GRU embedding: $\mathbf{\tilde{s_B}} \leftarrow \mathsf{GRU}(\mathbf{\hat{s_B}})$ 
        \STATE $Q, K, V \leftarrow \mathbf{h_A}, \mathbf{\tilde{s_B}} \oplus \mathbf{x_A}, \mathbf{\tilde{s_B}} \oplus \mathbf{x_A}$
        \STATE Unified standard probability score for context encoding $\mathbf{x_A}$: $\mathbf{p_{XA} = (1/x_A.dim[0]).repeat(x_A.dim[-1])}$
        \STATE $K_b \leftarrow \mathbf{p_B.repeat(\tilde{s_B}.dim[-1])} \oplus \mathbf{p_{XA}}$ 
        \STATE $\mathbf{\hat{h}_A} = \mathsf{Weighted Attention} (Q, K, V, K_b)$
        \\ \textit{// Joint Learning for Agent A Above}
        \STATE Coordinate alignment: $\mathbf{\hat{s_A}} \leftarrow \mathbf{s_A}$
        \STATE GRU embedding: $\mathbf{\tilde{s_A}} \leftarrow GRU(\mathbf{\hat{s_A}})$ 
        \STATE $Q, K, V \leftarrow \mathbf{h_B}, \mathbf{\tilde{s_A}} \oplus \mathbf{x_B}, \mathbf{\tilde{s_A}} \oplus \mathbf{x_B}$
        \STATE Unified standard probability score for context encoding $\mathbf{x_B}$: $\mathbf{p_{XB} = (1/x_B.dim[0]).repeat(x_B.dim[-1])}$
        \STATE $K_b \leftarrow \mathbf{p_A.repeat(\tilde{s_A}.dim[-1])} \oplus \mathbf{p_{XB}}$ 
        \STATE $\mathbf{\hat{h}_B} = \mathsf{Weighted Attention} (Q, K, V, K_b)$
        \\ \textit{// Joint Learning for Agent B Above}
        \IF {$Q > 1$}
        \STATE $\mathbf{h_A}$, $\mathbf{h_B} \leftarrow \mathbf{\hat{h}_A}, \mathbf{\hat{h}_B}$
        \STATE $\mathbf{(S_A, p_A)}$, $\mathbf{(S_B, p_B)} \leftarrow \mathsf{Marginal(h_A), Marginal(h_B)}$
        \ENDIF
        \ENDFOR 
    \RETURN $\mathbf{\hat{h}_A}$, $\mathbf{\hat{h}_B}$
    \end{algorithmic} 
\end{algorithm}

\subsection{Conditional Prediction\label{conditional}}
In this work, we use $\mathbf{S_A}$ and $\mathbf{S_B}$ to represent the set of multi-modal predictions for each agent, and $\mathbf{s_A}$ and $\mathbf{s_B}$ to represent an entry in the set. The problem of interactive motion prediction can be represented with conditional inference problem, as in \Cref{eq: cbpA}.

\begin{equation}\label{eq: cbpA}
    \mathbf{s^i_A} = p(\mathbf{s^i_A}| \mathbf{S_B}=\sum_{i=0}^N \mathbf{s^i_B}, \mathbf{x_A}), \quad \mathbf{s^i_A}\in \mathbf{S_A}
\end{equation}

Similarly, we apply the same notations to agent B and define the conditional inference of agent B as in \Cref{eq: cbpB}.
\begin{equation}\label{eq: cbpB}
\mathbf{s^i_B} = p(\mathbf{s^i_B}| \mathbf{S_A}=\sum_{i=0}^N \mathbf{s^i_A}, \mathbf{x_B}), \quad \mathbf{s^i_B}\in \mathbf{S_B}. 
\end{equation}

It is worth noting the conditional prediction is conditioned on all $N$ sampled trajectory candidates, with the pair-wise scoring module applied afterwards. \Cref{algo: jointlearning} listed the pseudo-code for the joint learning process. The joint learning module updates the agent embedding $\mathbf{h_A},\mathbf{h_B}$ with weight adjusted cross-attention mechanism, and we name it \textit{weighted attention}. $Q$ is the parameter determining the number of stacked \textit{weighted attention} to enhance interactive learning.

\begin{equation}\label{eq: weight_attention}
    \mathsf{WeightedAttention} = \frac{\mathsf{softmax}(Q^T K + Q^T K_b)}{\sqrt{d_k}}V
\end{equation}

In practice, the trajectory predictor makes predictions by sampling from the learned latent spaces $\mathbf{h_A},\mathbf{h_B} \in \mathbb{R}^{H\times E}$. The hidden spaces are \textit{heatmaps} \cite{gu2021densetnt} to represent the possible lanes the agent is following through an attention mechanism. The goal samples are further inferred through a probability estimation module. Finally, The predictor regresses the trajectory sequences corresponding to each selected goal. For agent A, the conditional inference is achieved by updating $\mathbf{h_A}$ with predicted trajectory candidates of agent B. That is, we are looking for $p(\mathbf{\hat{h_A}})$ in \Cref{eq: ha},

\begin{equation}\label{eq: ha}
    p(\mathbf{\hat{h_A}}) = p(\mathbf{h_A}|\mathbf{\tilde{s_B}}, \mathbf{x_A})
\end{equation}

where $\mathbf{\tilde{s_B}}$ is the trajectories $\mathbf{s_B} \in \mathbf{S_B}$ shifted to agent A's coordinate. We inference $p(\mathbf{\hat{h_A}}), p(\mathbf{\hat{h_B}})$ through \textit{weighted attention} as in \Cref{eq: weight_attention}. In \Cref{eq: weight_attention}, Q is the agent's embedding, K and V are aligned interactive agent's prediction embedding concatenated with feature encoding. $K_b$ is the mode probability score for each agent prediction concatenated with the normalized feature encoding score. Details are listed in \Cref{algo: jointlearning}. 

\subsection{Pair-Wise Scoring\label{socring}}
\begin{figure}[htbp]
    \centering
    \includegraphics[width=0.4\textwidth]{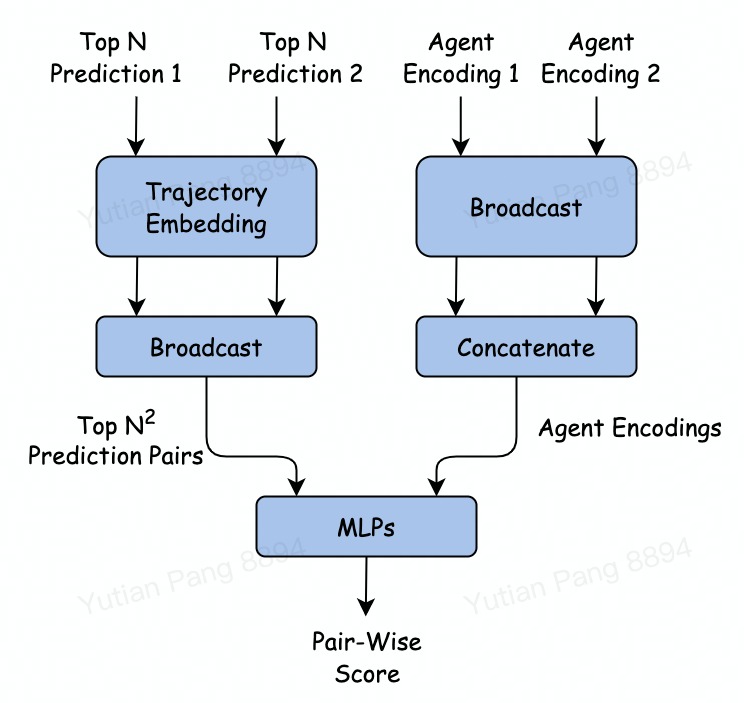}
    \caption{Pair-Wise Scoring: A simple two-layer MLP is adopted to output the pair-wise score. The MLP takes two inputs, the broadcasted top$N^2$ trajectory pairs, and each agent's encoding.}
    \label{fig: scorenet}
\end{figure}

The pair-wise scoring module is to select topK trajectory candidate pairs from sampled trajectories. \Cref{fig: scorenet} shows the structure of the pair-wise scoring module. The broadcasted top$N^2$ trajectory pairs ($\mathbf{s_A}, \mathbf{s_B}$) are concatenated with agent encoding ($\mathbf{\hat{h}_A}$, $\mathbf{\hat{h}_B}$) as the input to the two-layer MLPs ($\psi$). In trajectory scoring, we estimate the difference between ground truth and predictions for the entire sequence. We use the maximum entropy model to score topK trajectory pairs, similar to VectorNet but with extended dimensions. For the maximum entropy model, the loss is defined as the cross-entropy loss ($\Theta_{CE}$) between the predicted scores and the ground truth scores (\Cref{eq: score_loss}). 

\begin{equation}\label{eq: score_loss}
    \mathcal{L}_{score} := \Theta_{CE}(\psi(\mathbf{s_A}, \mathbf{\hat{h_A}}, \mathbf{s_B}, \mathbf{\hat{h_B}}), \mathcal{D}(\mathbf{s_A}, \mathbf{s_B}))
\end{equation}

Different to single motion prediction in VectorNet, we use averaged $l_\infty$ norm for trajectory pair in $\mathcal{D}$.

\begin{equation}\label{eq: l_norm}
    \mathcal{D}(\mathbf{s_A}, \mathbf{s_B}) = \frac{\mathsf{exp}(-(l_\infty(\mathbf{s},\mathbf{s^{GT}}))/2\alpha)}{\sum_{\mathbf{s'}}\mathsf{exp}((-l_\infty(\mathbf{s'},\mathbf{s^{GT}}))/2\alpha)}
\end{equation}

where $l_\infty(\mathbf{s},\mathbf{s^{GT}})$ is the averaged infinity norm of two interactive agents, 
\begin{equation}
    l_\infty(\mathbf{s},\mathbf{s^{GT}}) = \frac{1}{2} (l_\infty(\mathbf{s_A},\mathbf{s_A^{GT}}) + l_\infty(\mathbf{s_B},\mathbf{s_B^{GT}}))
\end{equation}

To determine the final topK trajectory candidate pairs after obtaining the trajectory score, we use the duplicate trajectory rejection mechanism. We first sort the trajectory candidate pairs based on their score, then find the next available candidate pair satisfying the separation threshold. We decay the threshold by a factor and repeat this process if there are no K pairs satisfying the threshold.

\section{Experiments\label{experiments}}
In this section, we introduce the procedure we performed to process the Waymo Motion Prediction Dataset for interactive prediction training as the feature engineering study of this research \Cref{features}. Finally, we show our interactive prediction results \Cref{results}.

\subsection{Dataset\label{dataset}}
We use Waymo Motion Prediction Data \cite{ettinger2021large} from the Waymo Open Motion Dataset (WOMD) to demonstrate the effectiveness of \name{}. WOMD is by far the largest public dataset intended for motion prediction research. It contains over 570 hours of driving data and spans over $1,750$km of roadways. The essential problem in WOMD is to predict the future 8 seconds positions given agents' 1-second histories sampled at 10Hz. WOMD provides test datasets for both single and interactive predictions and online data challenges for comparisons with state-of-the-art. However, the training set is kept the same for different prediction problems, with up to 8 prediction labels in each scene. Thus, the procedures for WOMD raw data processing and the determination of interactive pairs are the two fundamental problems for interactive behavior prediction. 

\subsection{Feature Engineering\label{features}}
Feature engineering is fundamental for deep learning practices. Good quality of data guarantees the success of the data-driven model, while a larger amount of data helps with generalizability. In this work, we perform feature processing from two aspects, (a) The quality of target samples. (b) The selection of interacting agents in the training set of Waymo Motion Prediction Dataset.

\noindent \textbf{Interaction Pairs}
In the WOMD training/validation set, the interactive pairs are not explicitly given. Thus we need to search at most $\mathsf{C}_8^2$ agent pairs for interactions. We use a heuristic-based method to determine interaction pairs in the WOMD training set. For any two predictable agents A and B in a scene, their ground truth trajectories are $\mathbf{s^{GT}_A}$ and $\mathbf{s^{GT}_A}$. We find the closest distance in the future locations of $\mathbf{s^{GT}_A}$ and $\mathbf{s^{GT}_A}$. If the closest $l_2$ distance in future timestamps is below the threshold (e.g., 5 meters in our case), we save the pair feature as the training set for interactive modeling (\Cref{eq: interaction}).

\begin{equation}\label{eq: interaction}
    d_{min} = \mathsf{min}_{t=10}^T \quad l_2(\mathbf{s^{GT}_A}, \mathbf{s^{GT}_B}) < 5
\end{equation}

\begin{figure}[htbp]
    \centering
    \includegraphics[width=0.45\textwidth]{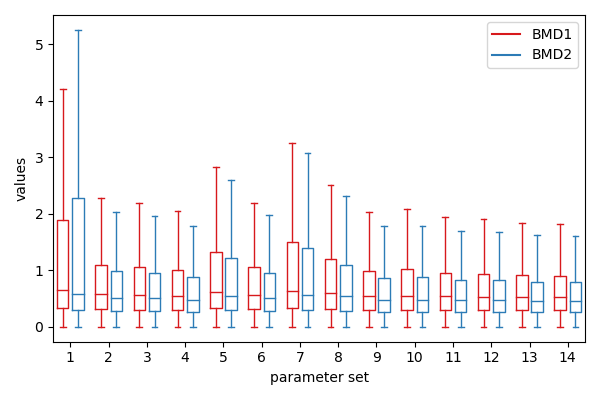}
    \caption{Boxplot of BMDs for 14 parameter sets in targets generation. Y-axis in meters.}
    \label{fig: targets distribution}
\end{figure}

\noindent \textbf{Target Samples} are the sampled destination points in target-driven predictors. Target driven method helps with enhancing multi-modality by performing regression conditioned on targets. Conversely, the training regressor suffers from mode averaging. In this work, targets are the sampled locations of $(x, y)$ an agent probably will reach at the last timestamp of $T$. As discussed in \Cref{conditional}, the trajectory predictor predicts a distribution of targets as location choices in the final timestamp. Under the assumption that the vehicles will not deviate far away from targets, we uniformly sample points based on the map centerlines. If we denote the offsets to centerlines as $\delta x$ and $\delta y$. Then the sampled targets are $\Gamma \in \mathbb{R}^N = \{\tau^\kappa\} = \{(x^\kappa, y^\kappa)+(\delta x^\kappa, \delta y^\kappa):\kappa \in \{1, 2,..., N\}\}$. In practice, we have several parameters controlling the sampling procedure: (a) The number of targets we want to sample. (b) The range of target samples (after coordinate alignment). (c) The range radius for searchable lanes. (d) The range radius for objects. For evaluation of target quality, we use the best mode displacement (BMD) on WOMD validation set, which is the minimum offsets of the sample targets. 
We can refer to Appendix. A. for detailed experiment records. As in \Cref{fig: targets distribution}, we select a parameter set with lower BMD for both interactive agents.  

\begin{figure*}
    \centering
    \begin{subfigure}[b]{0.475\textwidth}
        \centering
        \includegraphics[width=\textwidth]{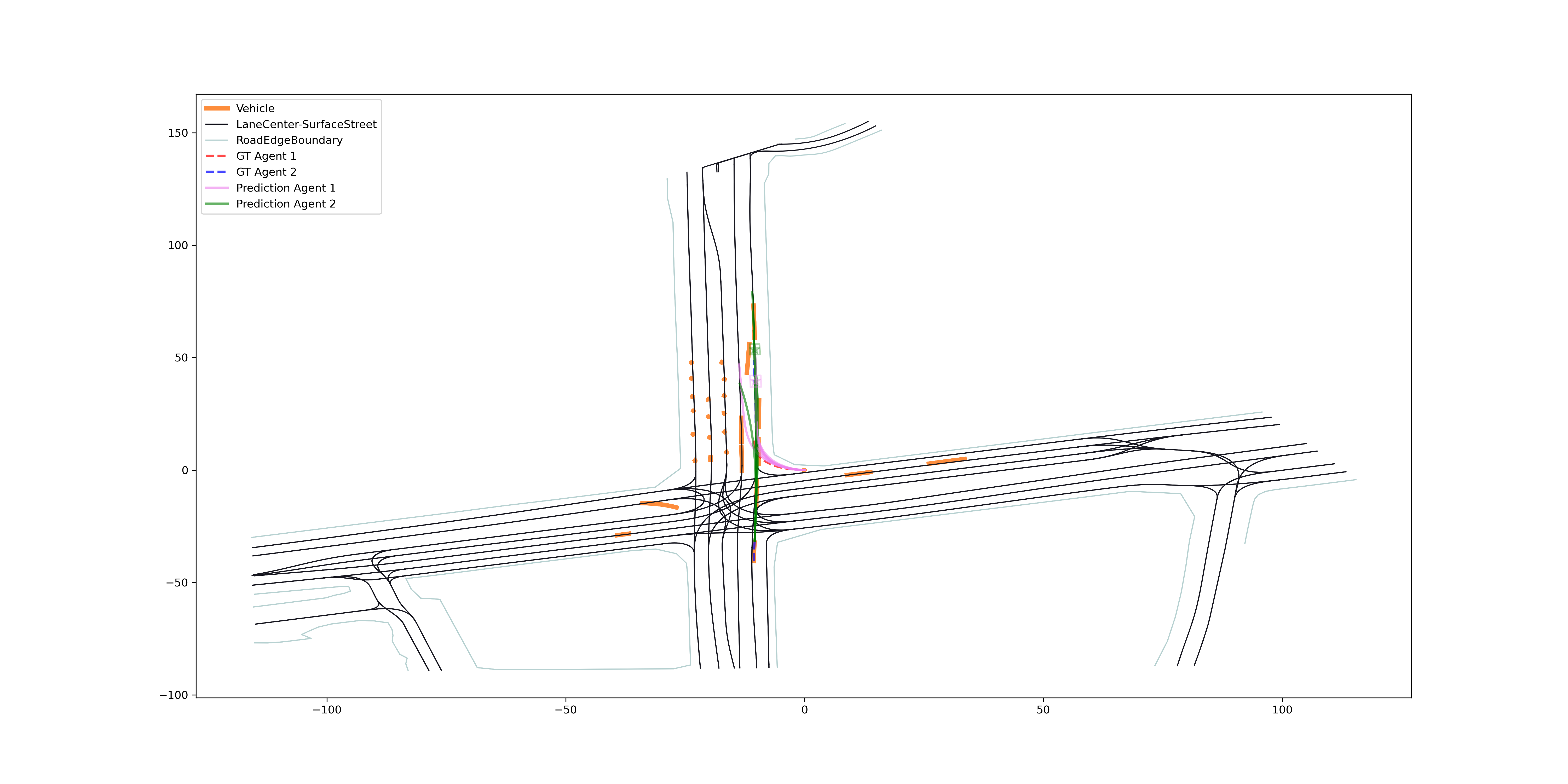}
        \caption[]
        {{\small Marginal Model Prediction for Scenario ID: 360b21f165487005}}    
        \label{fig: 360b21f165487005-marginal}
    \end{subfigure}
    \hfill
    \begin{subfigure}[b]{0.475\textwidth}  
        \centering 
        \includegraphics[width=\textwidth]{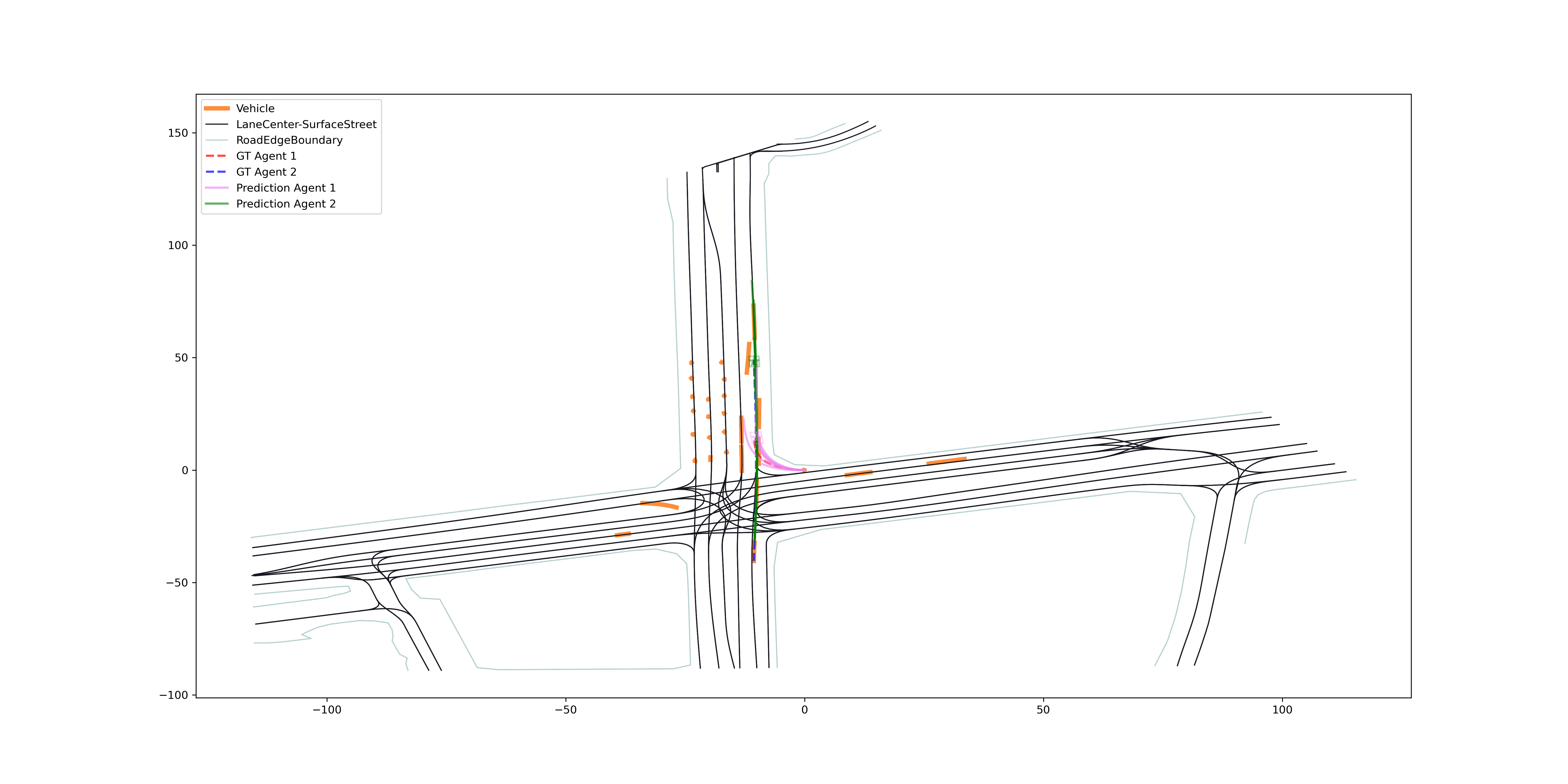}
        \caption[]%
        {{\small Joint Model Prediction for Scenario ID: 360b21f165487005}}    
        \label{fig: 360b21f165487005-joint}
    \end{subfigure}
    \vskip\baselineskip
    \begin{subfigure}[b]{0.475\textwidth}   
        \centering 
        \includegraphics[width=\textwidth]{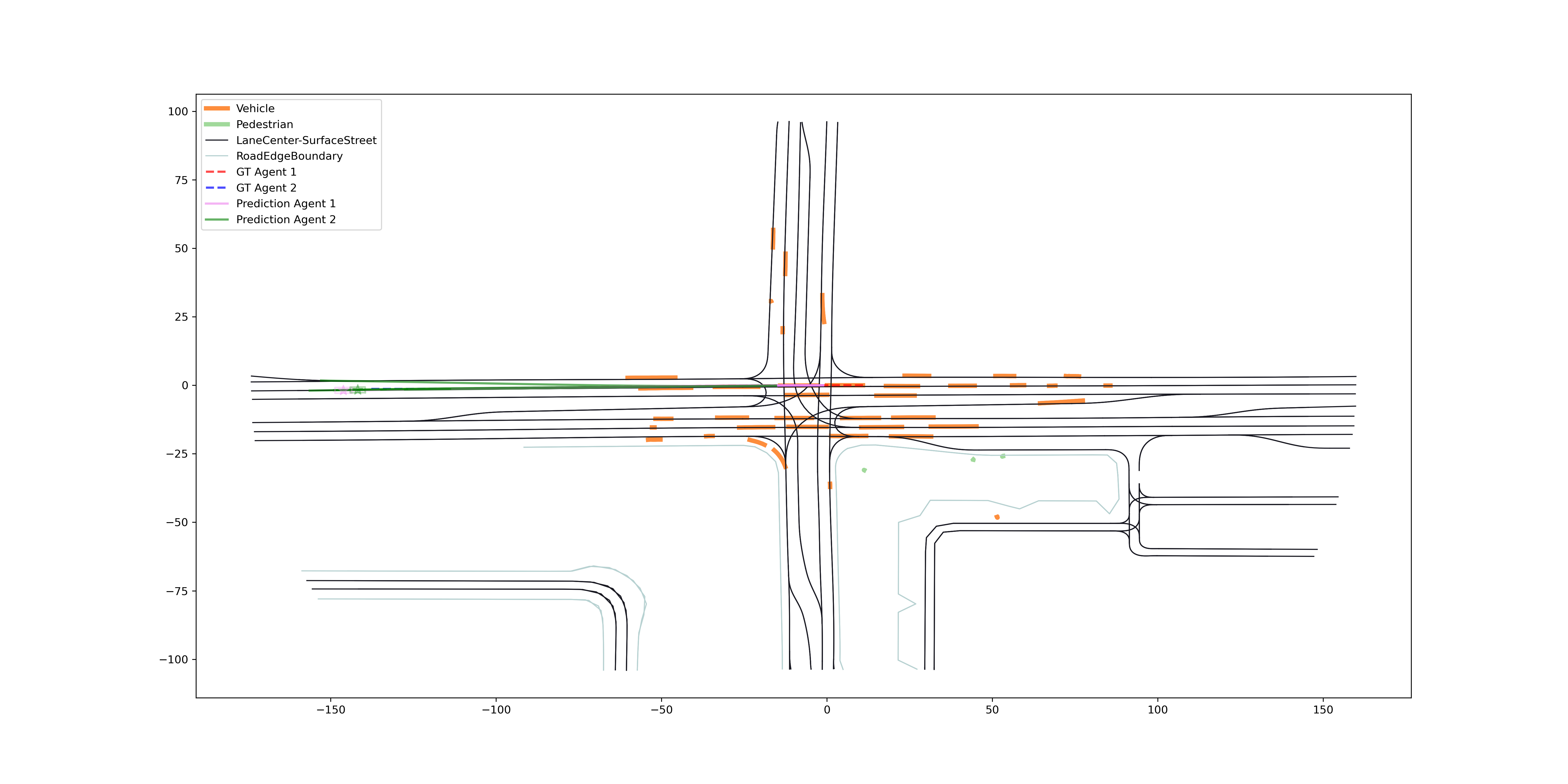}
        \caption[]%
        {{\small Marginal Model Prediction for Scenario ID: 8dd03cf09f4ae271}}    
        \label{fig: 8dd03cf09f4ae271-marginal}
    \end{subfigure}
    \hfill
    \begin{subfigure}[b]{0.475\textwidth}   
        \centering 
        \includegraphics[width=\textwidth]{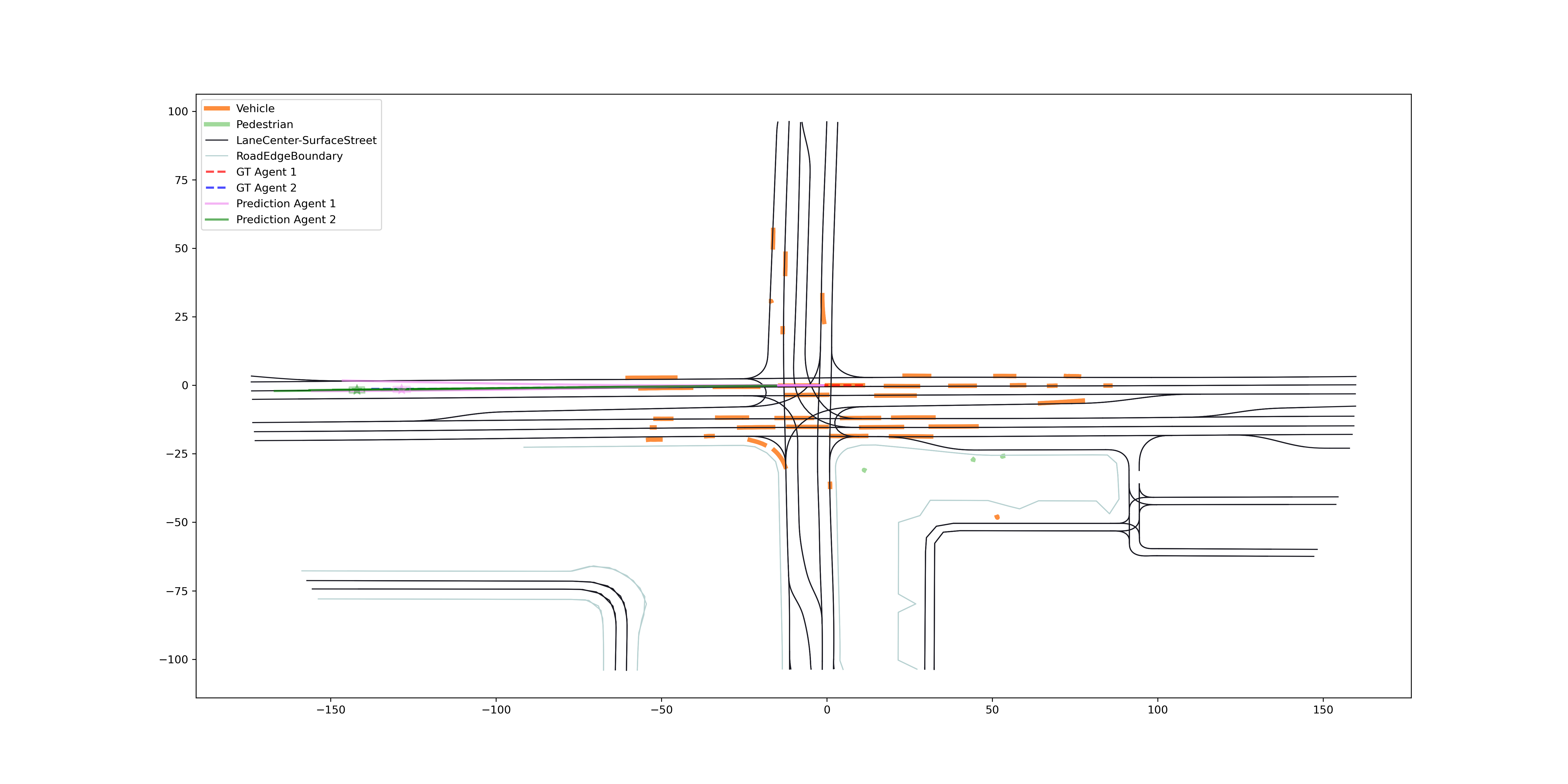}
        \caption[]%
        {{\small Joint Model Prediction for Scenario ID: 8dd03cf09f4ae271}}    
        \label{fig: 8dd03cf09f4ae271-joint}
    \end{subfigure}
    \vskip\baselineskip
    \begin{subfigure}[b]{0.475\textwidth}   
        \centering 
        \includegraphics[width=\textwidth]{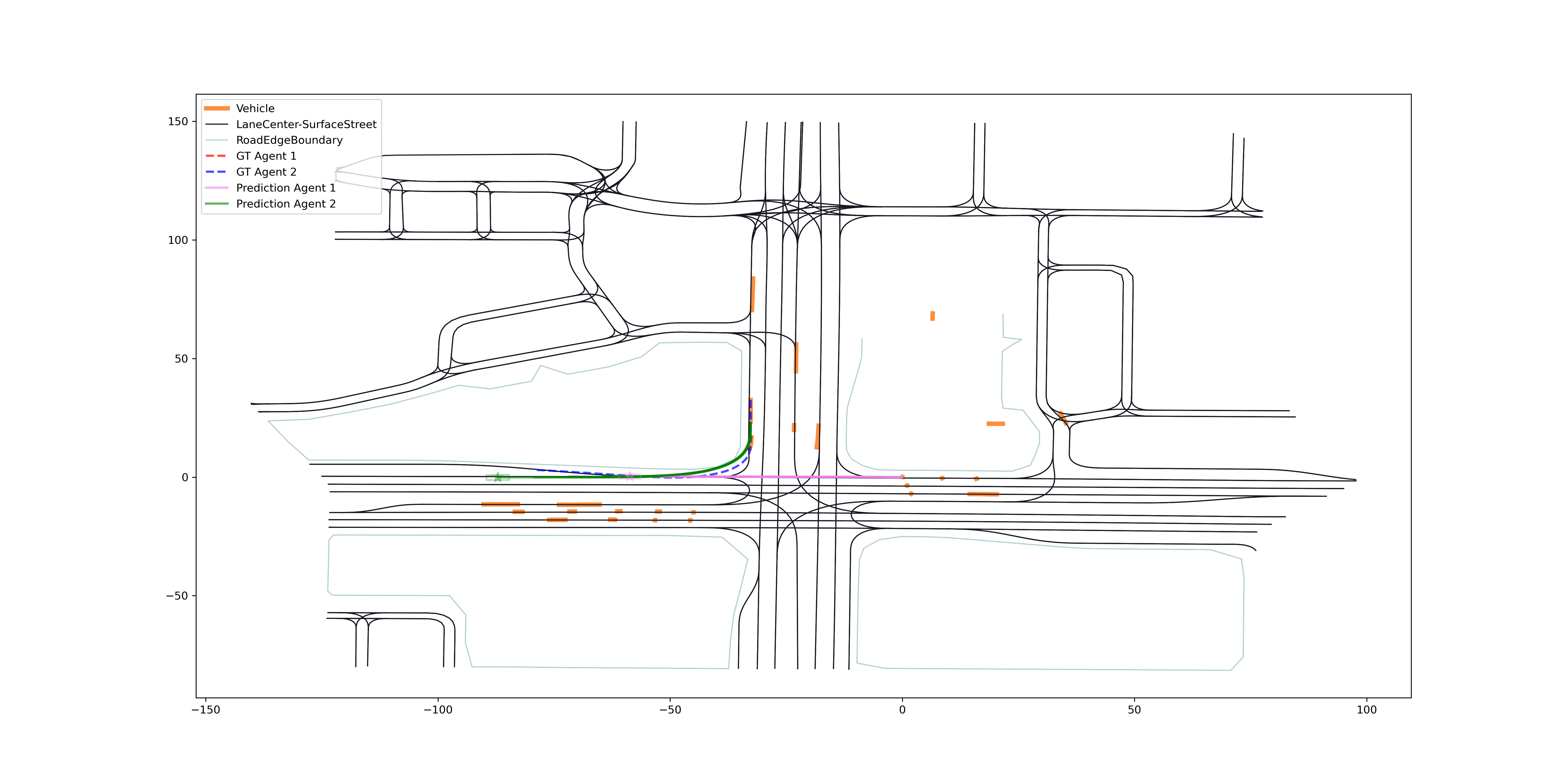}
        \caption[]%
        {{\small Marginal Model Prediction for Scenario ID: b1aece8720878be2}}    
        \label{fig: b1aece8720878be2-marginal}
    \end{subfigure}
    \hfill
    \begin{subfigure}[b]{0.475\textwidth}   
        \centering 
        \includegraphics[width=\textwidth]{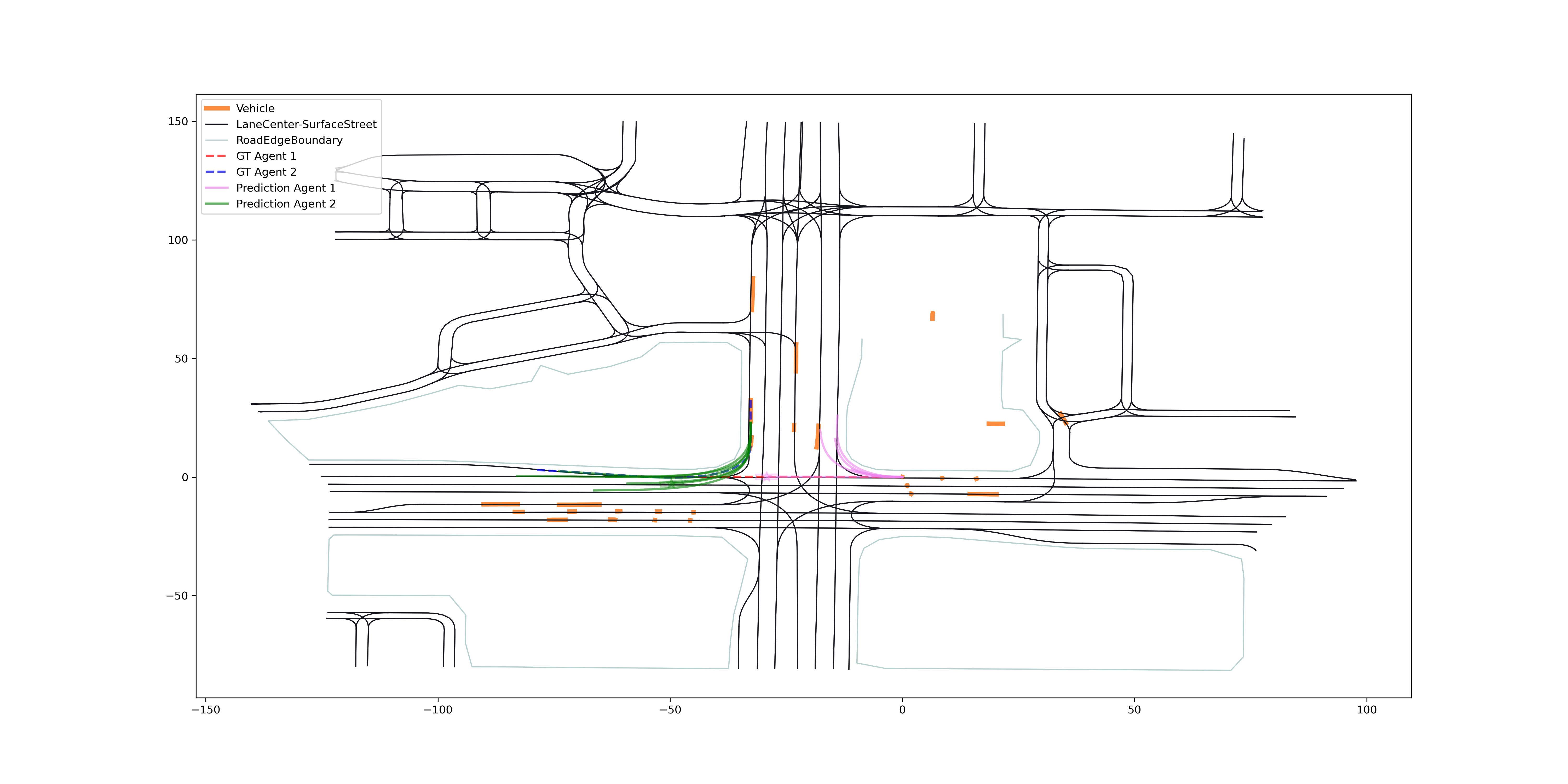}
        \caption[]%
        {{\small Joint Model Prediction for Scenario ID: b1aece8720878be2}}    
        \label{fig: b1aece8720878be2-joint}
    \end{subfigure}
    \caption[]
    {\small Visualization of corner cases. Left: marginal learning predictions. Right: joint learning predictions. Ground truths are in dashed lines. The multi-modal predictions of agents are marked in green and violet. Other nearby vehicles are in gold. The final timestamp of the best mode for both agents is marked with $\star$, with a bounding box representing vehicle shapes. In (a)$\&$(b), the right turn agent's joint prediction is yielding to the straight traffic, matching the ground truth. (c)$\&$(d) is the following scenario, the joint prediction of the following vehicle is keeping a safe distance to the leading vehicle, matching the ground truth. In (e)$\&$(f), the joint model forces agent A to have a right turn intent to avoid high interactions.} 
    \label{fig: visualizations}
\end{figure*}

\subsection{Evaluation Metrics\label{metrics}}
Similar to WOMD, we use the pair-wise evaluation metrics for multi-modal motion prediction. \textbf{minADE} and \textbf{minFDE} compute the minimum average/final displacement error of the best prediction mode. It is worth pointing out that the best mode in minADE is not necessary the best mode in minFDE, and vice versa. \textbf{OverlapRate} calculate the possible overlap between the predicting agent and other agents in the best mode by considering the Intersection over Union (IoU) for each agent with shapes. \textbf{MissRate} computes the possibility of neither mode making a correct prediction within a range of the ground truth, with speed adjustment to the range threshold. \textbf{Mean average precision (mAP)} is a general measure of the model performance by calculating the area under the precision-recall curve, where MissRate is used to define true positives. 

\subsection{Results\label{results}}

\begin{table}[htbp]
    \caption{Marginal Model v.s. Joint Model on the validation set} \label{table: marginaljoint}
    \begin{center}
        \resizebox{\columnwidth}{!}{
        \begin{tabular}{|c|c|c|c|c|c|}
            \hline
             & \textbf{minADE} & \textbf{minFDE} & \textbf{MissRate} & \textbf{OverlapRate} & \textbf{mAP} \\ \hline
            Marginal        & 4.499           & 12.596          & 0.816             & 0.378                & 0.045        \\ \hline
            Joint          & 3.012           & 8.118           & 0.826             & 0.416                & 0.115        \\ \hline
            \multicolumn{6}{l}{$^{\mathrm{a}}$The metrics are evaluated on the validation set.}\\
            \multicolumn{6}{l}{$^{\mathrm{b}}$The validation set only considers vehicles.}
        \end{tabular}}
    \end{center}
\end{table}

As shown in \Cref{fig: overview}(b), the joint learning model learns the interactions by updating the agent's embedding $\mathbf{h}$ through \textit{weighted attention}, and we name it \name{}. The updated agent embedding is further fed into the trajectory predictor to get paired predictions. To demonstrate the improved capability of our proposed method, we compare it with the marginal prediction baseline model in \Cref{fig: overview}(a). We adapt the marginal model for two agents by calculating the Cartesian Product of mode probabilities to get the \textit{naive} joint mode probability, and we get the topK predicted trajectories for interactive pairs. We compare the performance of \Cref{fig: overview}(a) and \Cref{fig: overview}(b) and list the results in \Cref{table: marginaljoint}. For this comparison, we first fine-tune the marginal model towards single motion prediction state-of-the-art, then use the same model parameters to train the joint model for a fair comparison.

\begin{table}[htbp]
\caption{Comparison with other methods.} \label{table: jointsota}
\begin{center}
    \resizebox{\columnwidth}{!}{
    \begin{tabular}{|c|c|c|c|c|c|}
        \hline
        \textit{}                                                       & \textbf{minADE} & \textbf{minFDE} & \textbf{MissRate} & \textbf{OverlapRate} & \textbf{mAP}  \\ \hline
        \multicolumn{6}{|c|}{\textit{\textbf{Validation Set}}}                                                                                                         \\ \hline
        \begin{tabular}[c]{@{}c@{}}Scene\\ Transformer\end{tabular}     & 1.72            & 3.99            & 0.49              & -                    & 0.11          \\ \hline
        M2I                                                             & -               & 5.49            & 0.55              & -                    & 0.18          \\ \hline
        \textit{\name{}}                                                    & \textit{3.01}   & \textit{8.12}   & \textit{0.83}     & -       & \textit{0.12} \\ \hline
        \multicolumn{6}{|c|}{\textit{\textbf{Test Set}}}                                                                                                               \\ \hline
        \begin{tabular}[c]{@{}c@{}}Scene\\ Transformer\end{tabular}     & 1.76            & 4.08            & 0.50              & 0.28                 & 0.10          \\ \hline
        M2I                                                             & 2.50            & 5.65            & 0.57              & 0.39                 & 0.16          \\ \hline
        HeatRM4                                                         & 2.93            & 7.20            & 0.80              & 0.46                 & 0.07          \\ \hline
        \begin{tabular}[c]{@{}c@{}}Waymo\\ LSTM\\ Baseline\end{tabular} & 4.52            & 12.40           & 0.87              & 0.56                 & 0.01          \\ \hline
        \multicolumn{6}{l}{$^{\mathrm{a}}$The results here only consider vehicles.}\\
        \multicolumn{6}{l}{$^{\mathrm{b}}$The test set results are obtained from motion challenge leaderboard.}
    \end{tabular}}
\end{center}
\end{table}

We present the statistical comparison of \name{} with other SOTA models on interactive motion prediction in \Cref{table: jointsota}. We show that \name{} achieves comparable performance against other SOTA benchmarks. Additionally, \name{} appears to have superior prediction capability, even surpassing SceneTransformer, in terms of mAP. A large portion of the benefits come from the conditional inference process. Additionally, we visualize several scenes as corner cases to demonstrate the interaction-aware prediction in \Cref{fig: visualizations}. We show that the joint learning process can improve high interaction predictions, especially in yielding/following scenarios. In some scenarios, the joint model may give \textit{over-conservative} predictions to reduce potential conflict. These types of behaviors may not always match the ground truth but are reasonable, as in \Cref{fig: visualizations}(e)$\&$(f). More corner cases are included in Appendix B. 

\section{Discussions\label{discussions}}

The proposed end-to-end \name{} model can be extended to multi-agents interactive prediction. The joint learning block in \Cref{fig: overview} can be generalized to conditional encoding of multiple agents by extending the attention layer dimension in \Cref{algo: jointlearning}. In \Cref{algo: jointlearning}, the key-value tensor pairs will be expand to include the predicted trajectories of all interacting agents. For three interacting agents ({$A,B,C$}), during updating the encoding of agent,  $K, V \leftarrow \mathbf{\tilde{s_B}} \oplus \mathbf{\tilde{s_C}} \oplus \mathbf{x_A}$. Similarly, the biased key tensor pairs need to concatenate predicted trajectory probabilities to match the dimension in \Cref{eq: weight_attention}. Also, the scoring function can be straightly adjusted to multi-agents candidates selection function. However, the proposed joint scoring function has exponential complexity increase in the broadcast phase of \Cref{fig: jointblock}. The experiments of multiple interacting agents are beyond the scope of this paper, and the WOMD behavior prediction dataset also limit the problem setup to interactive agent pairs. We leave them as the major studies to investigate in the future.

Different than M2I \cite{sun2022m2i}, we treat each agent in the interactive pairs equally. We believe the interactive agents are equally interacted even if there are clearly defined right-of-ways. When building real-world safety critical applications, we should think about the problem from practical perspective. The agent who has the right-of-ways should always pay attention to nearby agents on the road, in case something unexpected happened (and it's very likely to encounter reckless driver on the road). In real-world, we shouldn't make the assumption that everyone follows the moral standards and behave as what's required, and it's very not likely to be happen. Regarding the model itself, we treat each agent equally interacted doesn't mean the learned policy treat each agent equally. Our proposed module try to perform cross attention on each agent's embedding space to force them be aware of each other. If the traffic rules are retained within the real-world data, the model is certainly able to capture them. For example, the multi-modal prediction results for the agent with right-of-ways could output predictions based on the regulations, or trying to avoid the other interacting agent for safety assurance (with a smaller mode probability). With manually defined influencer-reactor relationships, the prediction won't consider alternative routes but hold the current right-of-ways. This intuition also guides the development of the model architecture. For instance, the evaluation of interactive trajectory pair predictions has equal weight towards the calculation of pair-wised prediction infinity norms. The usage of 2-layer MLP and $\psi$ follow the original target-driven prediction paper (TNT) where a detailed explanation is given.

\subsection{Insights\label{insights}}
Our method has limitations and can be further improved in many aspects. We give a few potential research directions here.

\begin{itemize}
    \item We are expecting a performance improvement by fine-tuning the parameters of the interactive model. Currently, we are using the same model parameters for both the marginal model and joint model to show the effectiveness of the proposed joint learning block. However, to achieve a better performance, we can adjust the embedding dimensions, the number of stacked \textit{weighted attention} modules, the learning schedulers, etc. By performing an extensive parametric study, the joint model performance can be improved. 
    \item During the feature processing, we are using the heuristic-based method to select an interacting agent within the training dataset. The heuristic largely impacted the quality of the training set of the model, while the training set is the most critical factor towards a good model. We propose to try different feature selection metrics for the preparation of the training data, for instance, \textit{mutual information} or \textit{degree of influence} \cite{tolstaya2021identifying,ettinger2021large}. 
    \item Interactive trajectory prediction is a less-explored area. The literature on interactive trajectory prediction assumes there are only two interacting agents on operation. Multi-agent interactions or dynamic changing of interactive agents scenarios would definitely be the future research directions to the community.
    \item There are potential improvements in the pair-wise scoring module. For instance, we can perform a parametric study on the selection of norms on different timestamps. We can alternate the $l_\infty$ norm to the Euclidean norm, or $l_0$ norm, and adjust the timestamps to calculate the difference. 
    \item The trajectory selection function of non-maximum suppression can be further investigated. At this point, we can apply different speed penalized selection thresholds on the driving direction and its tangent direction to reflect the impact of the vehicle speed profile.
    \item We haven't looked into alternative trajectory predictors to study the generalizability of the proposed joint learning framework. Similar to this work, \name{} can be applied to the latent space for any marginal predictor.
\end{itemize}

\section{Conclusions\label{conclusions}}
In summary, we propose an interactive motion prediction model, \name{}. For interaction-aware prediction, the realization of conditional inference is crucial. In \name{}, we perform conditional inference within the joint learning block, where we propose a novel \textit{weighted attention} mechanism to update the agent's trajectory latent spaces from the marginal learning model. The updated agent representation for both agents is aware of all the topN predicted future positions of the interacting agent. Consequently, \name{} outputs interaction-aware of the sampled trajectories. We show that our proposed model achieves performance improvement by comparing it with a single motion prediction model through experiments on the Waymo Motion Prediction Dataset. The high mAP without extensive parameters tuning demonstrates that \name{} is good at capturing interactions through explicit interaction modeling, which gives insights for future studies. Additionally, we show that \name{} achieves comparable performance to the state-of-the-art. Furthermore, we visualize the predictions for several scenarios in the test set. These visualizations demonstrate the effectiveness of our proposed learning framework.

\section*{Acknowledgment}
The authors thank the support from Xsense.ai Inc.

\bibliographystyle{IEEEtran}
\bibliography{ref.bib}
\vfill
\pagebreak
\onecolumn
\section*{Appendix\label{appendix}}
\subsection*{A. Target Generation Parameters\label{targets}}

This is to list the parametric study mentioned during the feature preparation process, as discussed in Sec IV-B.
\begin{table}[htbp]
\caption{Experiment Records for Target Samples}\label{table: targets}
\begin{adjustbox}{width=\columnwidth,center}
    \begin{tabular}{c|c|c|c|c|c|c|c||c|c|c|c|c}
        \hline
        \textbf{Parameter Set Number} & \textbf{Number of Target} & \textbf{Range of Target}                    & \textbf{Radius of Lane} & \textbf{Radius of Object} & \textbf{BMD1:Mean} & \textbf{BMD1:Median} & \textbf{BMD1:75\%} & \textbf{BMD1:90\%} & \textbf{BMD2:Mean} & \textbf{BMD2:Median} & \textbf{BMD2:75\%} & \textbf{BMD2:90\%} \\ \hline
        \#1                           & 8000                      & {[}-100.0, 50.0, -80.0, 80.0{]}             & 80                      & 60                        & 9.87               & 0.65                 & 1.89               & 40.60              & 9.54               & 0.58                 & 2.29               & 38.73              \\ \hline
        \#2                           & 8000                      & {[}-150.0, 100.0, -100.0, 100.0{]}          & 160                     & 120                       & 3.30               & 0.58                 & 1.10               & 3.28               & 3.02               & 0.51                 & 0.98               & 2.71               \\ \hline
        \#3                           & 8000                      & {[}-200.0, 100.0, -150.0, 150.0{]}          & 160                     & 120                       & 3.05               & 0.57                 & 1.07               & 2.36               & 2.81               & 0.50                 & 0.95               & 2.00               \\ \hline
        \textbf{\#4}                  & \textbf{10000}            & \textbf{{[}-200.0, 100.0, -150.0, 150.0{]}} & \textbf{160}            & \textbf{120}              & \textbf{2.24}      & \textbf{0.55}        & \textbf{1.00}      & \textbf{1.84}      & \textbf{1.96}      & \textbf{0.48}        & \textbf{0.88}      & \textbf{1.63}      \\ \hline
        \#5                           & 6000                      & {[}-150.0, 100.0, -100.0, 100.0{]}          & 120                     & 100                       & 6.00               & 0.62                 & 1.33               & 21.29              & 5.69               & 0.55                 & 1.22               & 21.26              \\ \hline
        \#6                           & 8000                      & {[}-300.0, 200.0, -200.0, 200.0{]}          & 200                     & 200                       & 2.90               & 0.57                 & 1.06               & 2.11               & 2.73               & 0.51                 & 0.96               & 1.98               \\ \hline
        \#7                           & 6000                      & {[}-200.0, 100.0, -150.0, 150.0{]}          & 100                     & 100                       & 7.58               & 0.63                 & 1.50               & 30.22              & 7.24               & 0.56                 & 1.41               & 28.20              \\ \hline
        \#8                           & 6000                      & {[}-200.0, 100.0, -150.0, 150.0{]}          & 160                     & 120                       & 4.45               & 0.60                 & 1.20               & 8.68               & 4.31               & 0.54                 & 1.10               & 9.26               \\ \hline
        \#9                           & 10000                     & {[}-200.0, 100.0, -100.0, 100.0{]}          & 160                     & 120                       & 2.21               & 0.55                 & 1.00               & 1.83               & 1.94               & 0.48                 & 0.88               & 1.61               \\ \hline
        \#10                          & 12000                     & {[}-200.0, 100.0, -150.0, 150.0{]}          & 140                     & 100                       & 2.52               & 0.55                 & 1.02               & 2.04               & 2.12               & 0.48                 & 0.88               & 1.74               \\ \hline
        \#11                          & 12000                     & {[}-200.0, 100.0, -100.0, 100.0{]}          & 160                     & 120                       & 1.73               & 0.54                 & 0.96               & 1.67               & 1.47               & 0.47                 & 0.84               & 1.47               \\ \hline
        \#12                          & 12000                     & {[}-200.0, 100.0, -100.0, 100.0{]}          & 200                     & 200                       & 1.38               & 0.54                 & 0.94               & 1.56               & 1.23               & 0.47                 & 0.83               & 1.42               \\ \hline
        \#13                          & 16000                     & {[}-200.0, 100.0, -100.0, 100.0{]}          & 200                     & 200                       & 0.99               & 0.53                 & 0.91               & 1.47               & 0.87               & 0.46                 & 0.80               & 1.34               \\ \hline
        \#14                          & 20000                     & {[}-200.0, 100.0, -100.0, 100.0{]}          & 200                     & 200                       & 0.94               & 0.52                 & 0.91               & 1.47               & 0.81               & 0.46                 & 0.80               & 1.33               \\ \hline
        
    \end{tabular}
\end{adjustbox}
\end{table}

\subsection*{B. Interaction Scenarios Visualization \label{figures}}
In this section, we visualize several scenarios in Waymo Interactive Motion Prediction Dataset. All lines are rotated back to agent A's coordinate system. The line legends can be found in the upper-left corner. For abbreviation, we use \textit{the marginal model} to stand for the Cartesian product of two single motion prediction model outputs and use the \textit{joint model} to represent the interactive motion prediction model.

\begin{itemize}
    \item For marginal model visualization, each agent's top 6 trajectory candidates are drawn, with the corresponding vehicle shape visualized on the destination of the highest probability mode ($T_{pred}=8s$). 
    \item For joint model visualization, top 6 trajectory candidate pairs are drawn, with vehicle shape visualized on the destination of the highest probability pairs ($T_{pred}=8s$).
\end{itemize}

Moreover, we separate these visualizations into different categories for classified illustrations.\\
\textbf{Joint model matches the ground truth and shows higher prediction capability.}

\begin{figure*}[ht!]
    \centering
    \begin{subfigure}[b]{0.475\textwidth}
      \includegraphics[width=\linewidth]{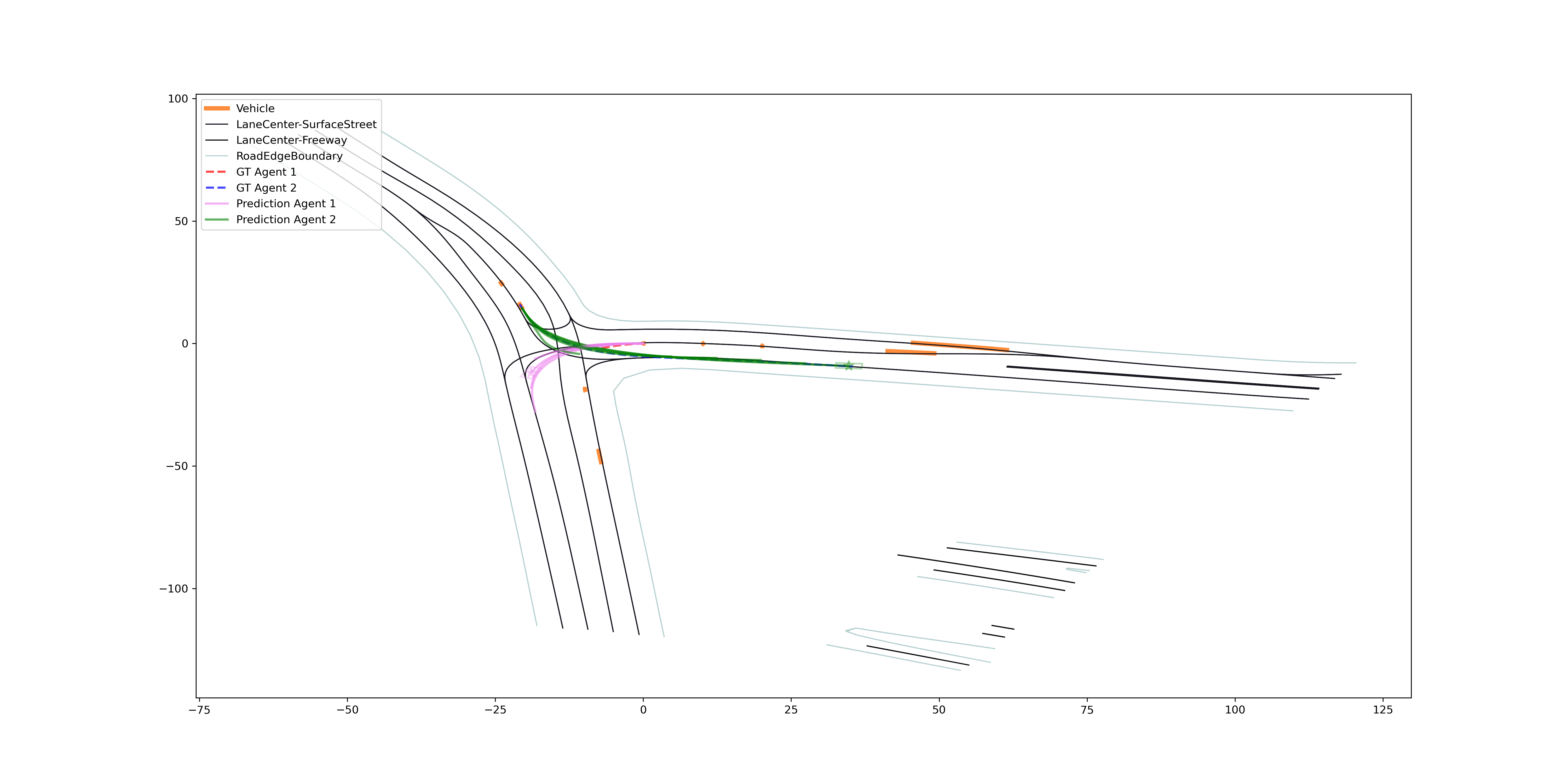}
      \caption{\small Marginal Model Prediction for Scenario ID: b4dfa76ba3ab3b02}
    \end{subfigure}
    \hfill
    \begin{subfigure}[b]{0.475\textwidth}
      \includegraphics[width=\linewidth]{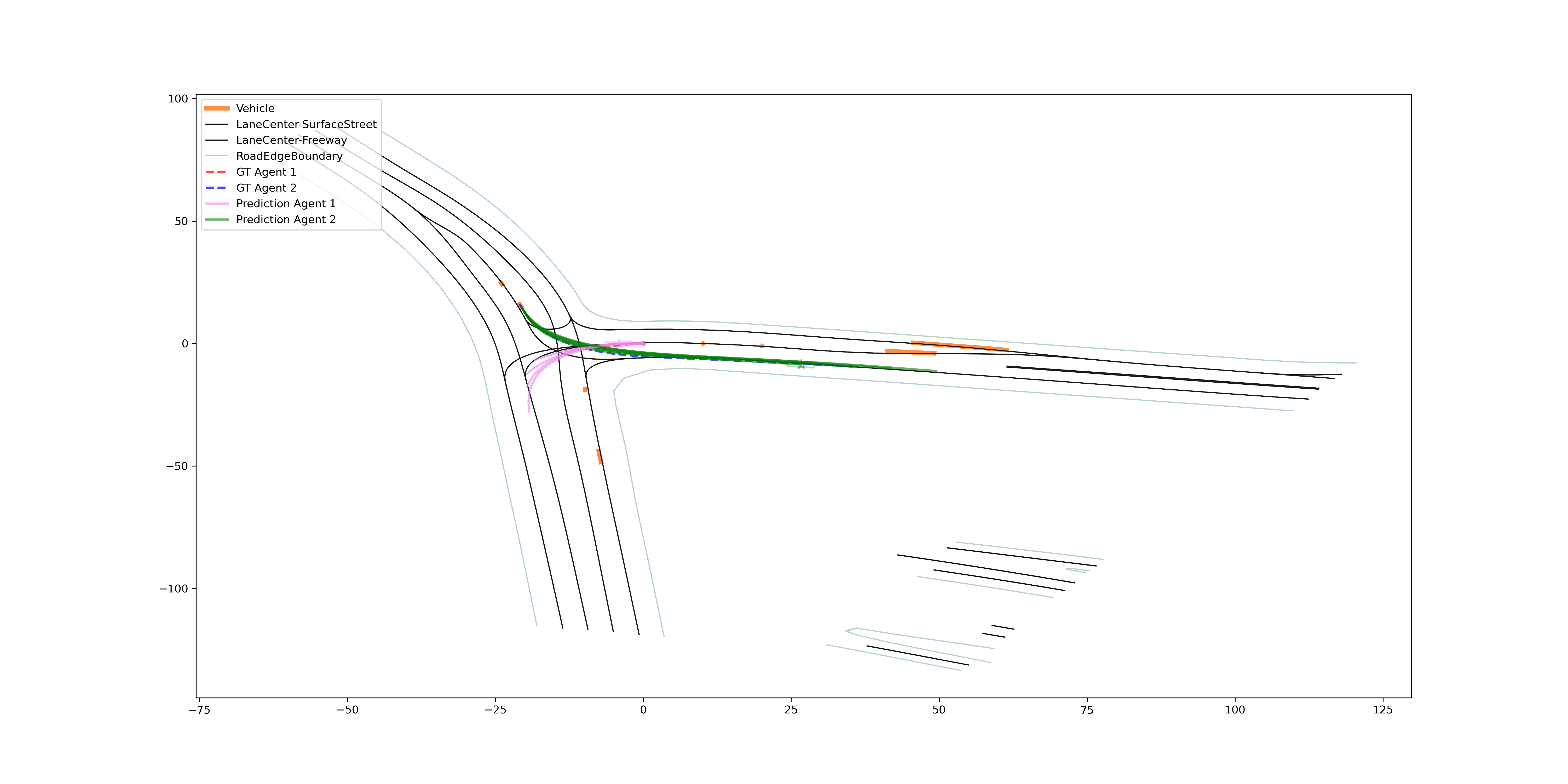}
      \caption{\small Joint Model Prediction for Scenario ID: b4dfa76ba3ab3b02}
    \end{subfigure}
    \caption{\small During a highly interactive two-left-turn scenario, the joint prediction of agent A yields to agent B, matching the ground truth.}
\end{figure*}

\begin{figure*}[ht!]
    \centering
    \begin{subfigure}[b]{0.475\textwidth}
      \includegraphics[width=1\linewidth]{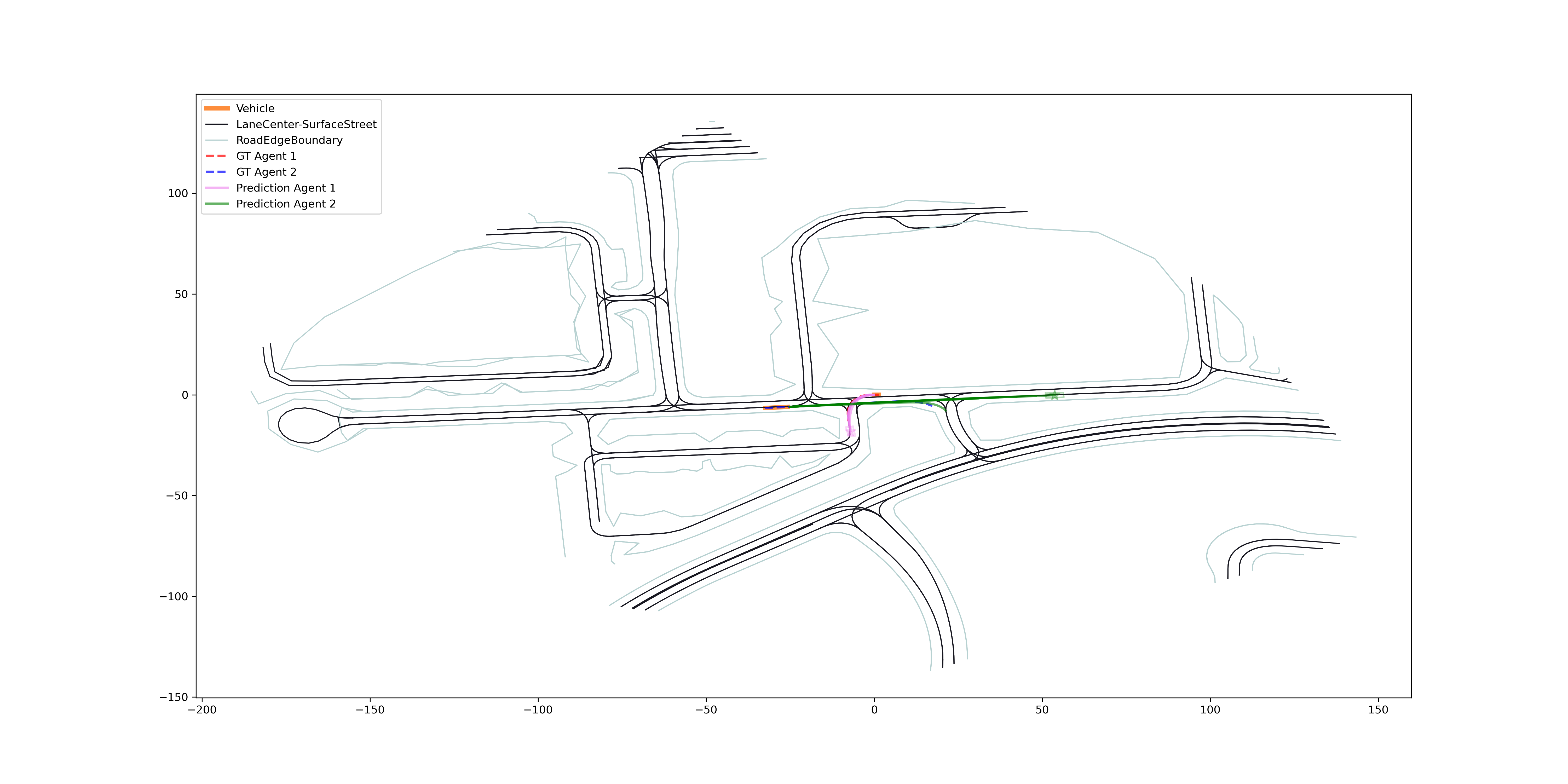}
      \caption{\small Marginal Model Prediction for Scenario: cda124af5393dff4}
    \end{subfigure}
    \hfill
    \begin{subfigure}[b]{0.475\textwidth}
      \includegraphics[width=1\linewidth]{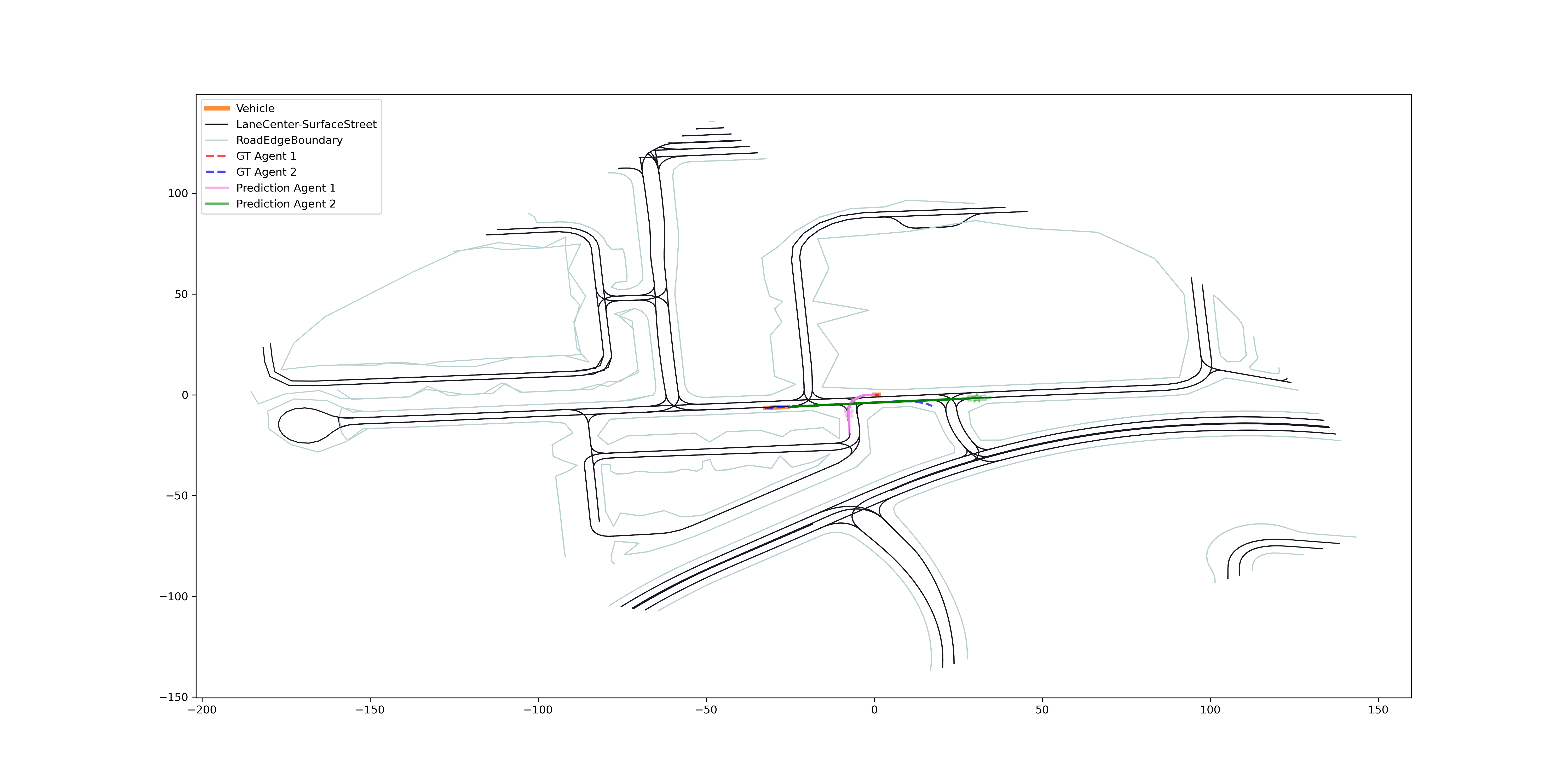}
      \caption{\small Joint Model Prediction for Scenario ID: cda124af5393dff4}
    \end{subfigure}
    \caption{\small Agent A's joint prediction is yielding to the straight traffic by adjusting speed, matching the ground truth.}
\end{figure*}

\begin{figure*}[ht!]
    \centering
    \begin{subfigure}[b]{0.475\textwidth}
      \includegraphics[width=1\linewidth]{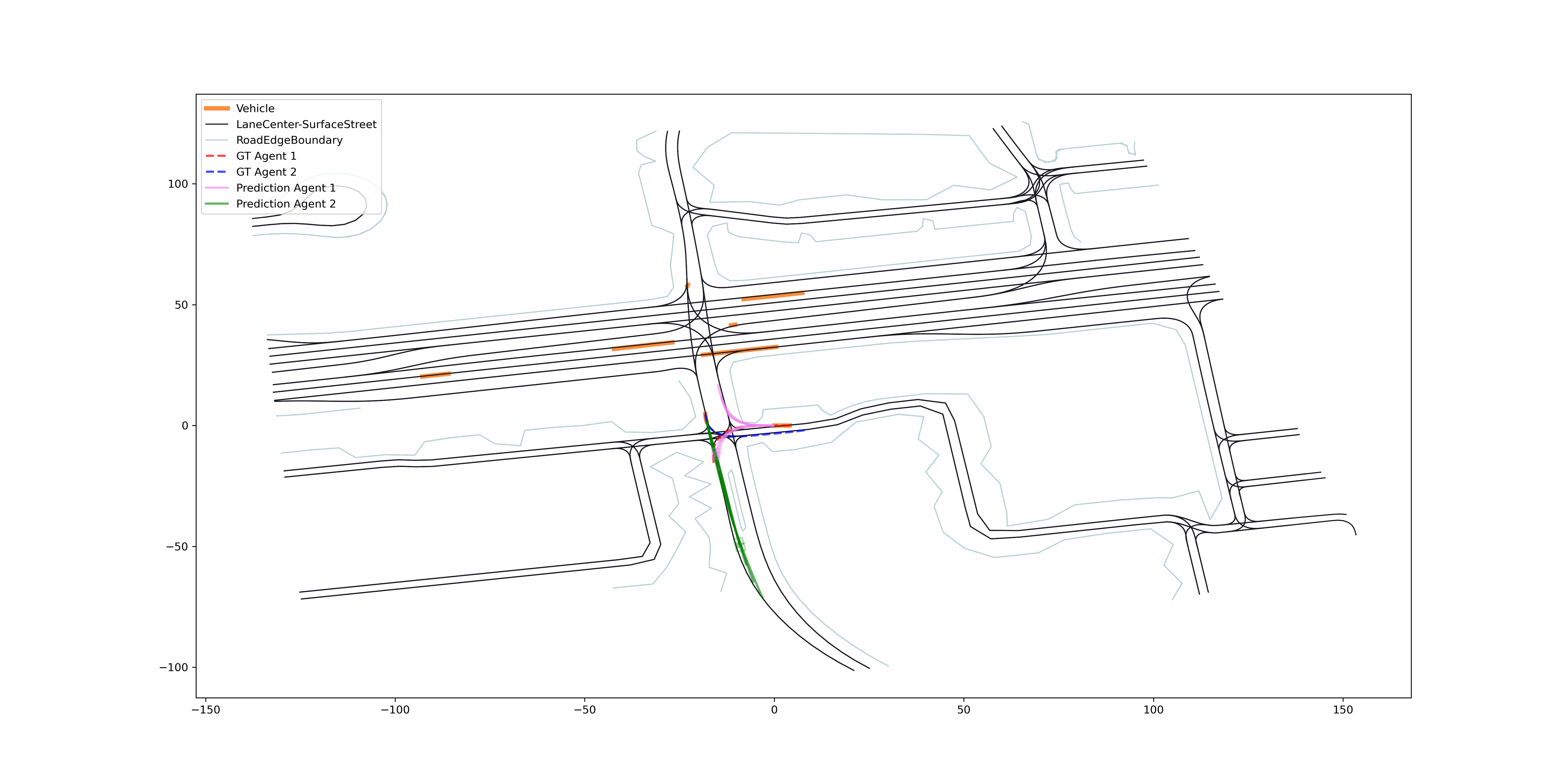}
      \caption{\small Marginal Model Prediction for Scenario: 6b011fa4a87377e3}
    \end{subfigure}
    \hfill
    \begin{subfigure}[b]{0.475\textwidth}
      \includegraphics[width=1\linewidth]{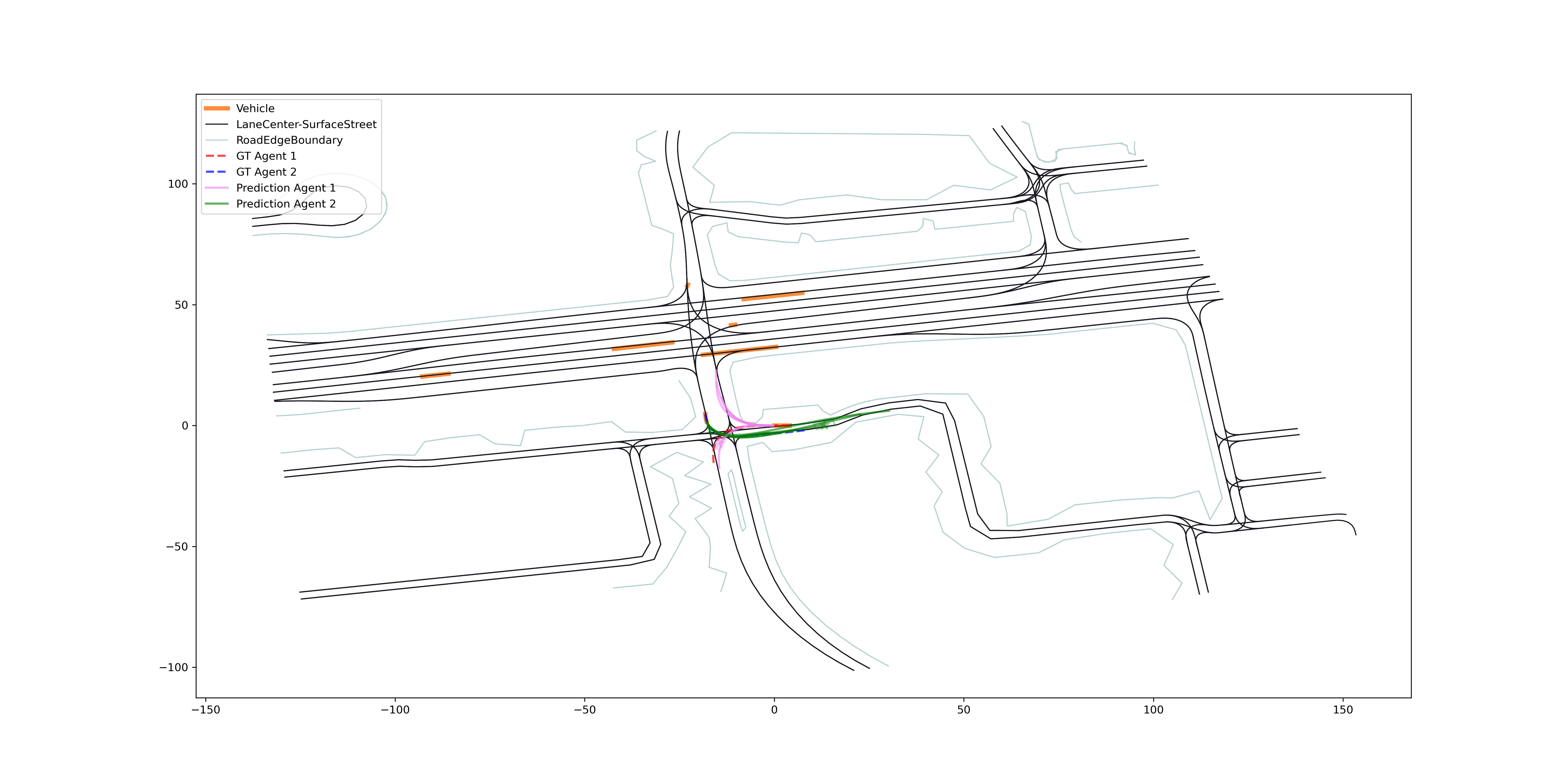}
      \caption{\small Joint Model Prediction for Scenario ID: 6b011fa4a87377e3}
    \end{subfigure}
    \caption{\small The joint prediction model is aware of the two-left-turns scenario and matches the ground truth}
\end{figure*}

\begin{figure*}[ht!]
    \centering
    \begin{subfigure}[b]{0.475\textwidth}
      \includegraphics[width=1\linewidth]{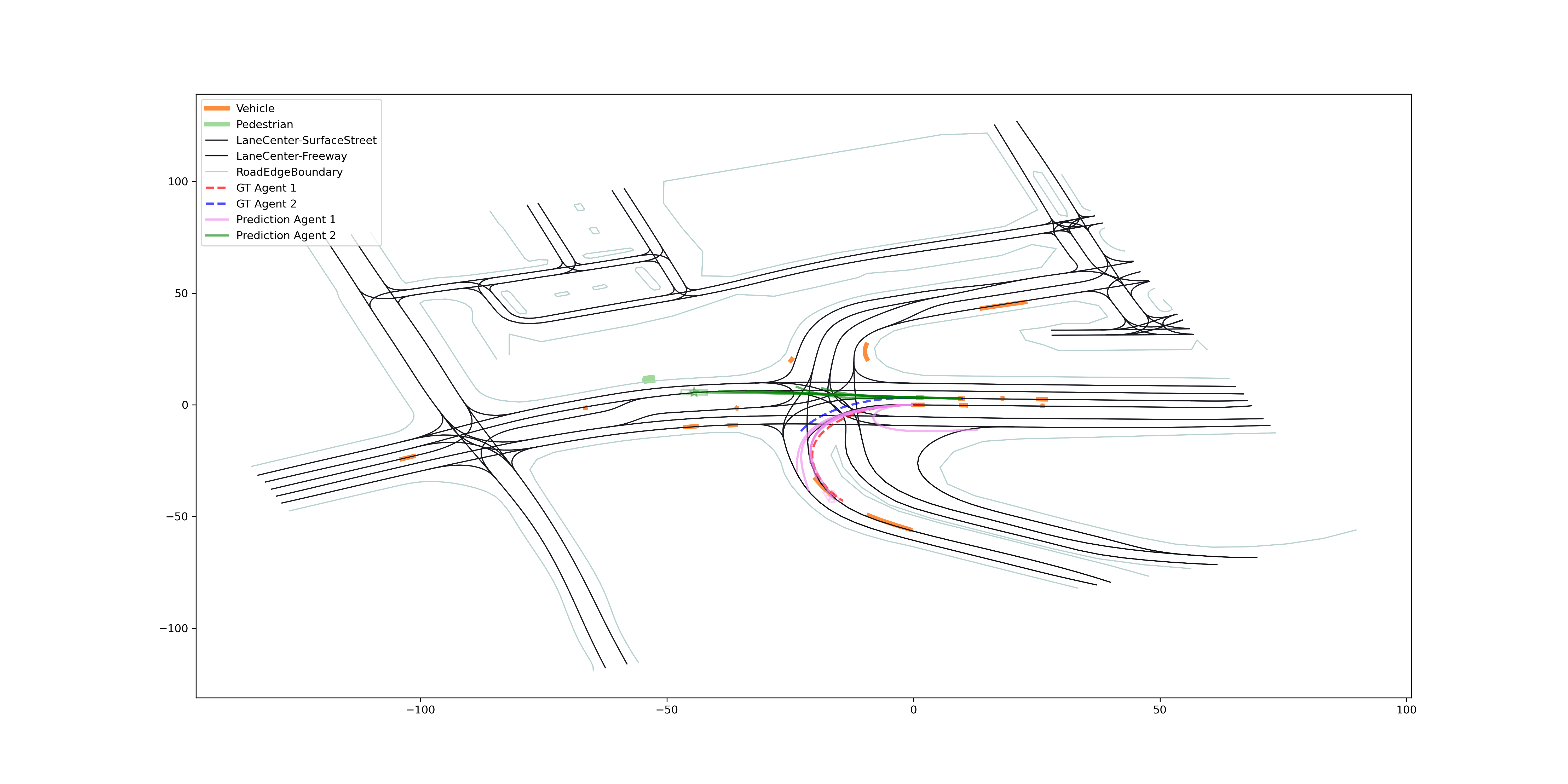}
      \caption{\small Marginal Model Prediction for Scenario: 4389c3b5fd11d3b7}
    \end{subfigure}
    \hfill
    \begin{subfigure}[b]{0.475\textwidth}
      \includegraphics[width=1\linewidth]{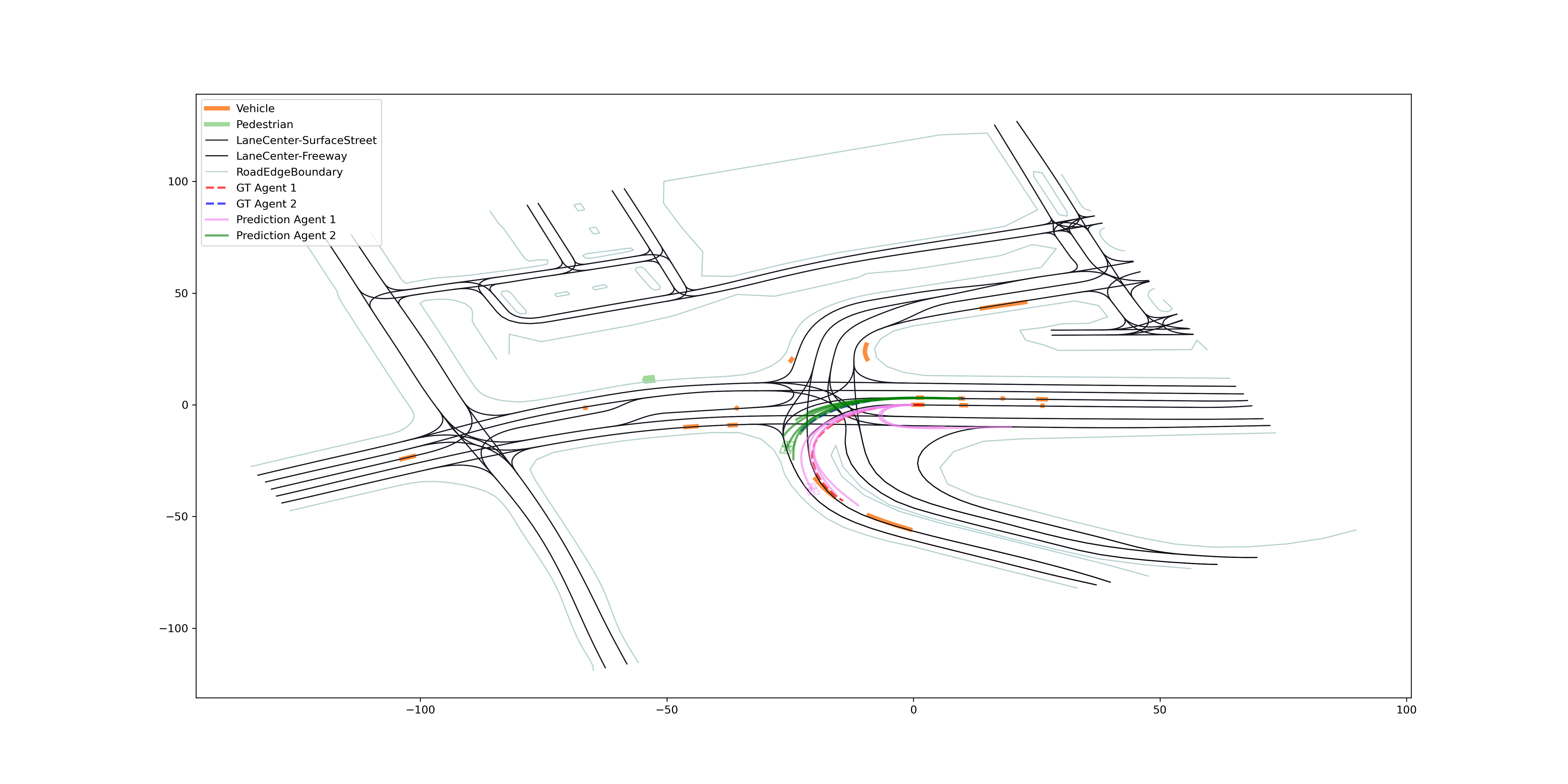}
      \caption{\small Joint Model Prediction for Scenario ID: 4389c3b5fd11d3b7}
    \end{subfigure}
    \caption{\small The joint prediction model is aware of the two-left-turns scenario and matches the ground truth}
\end{figure*}

\begin{figure*}[ht!]
    \centering
    \begin{subfigure}[b]{0.475\textwidth}
      \includegraphics[width=1\linewidth]{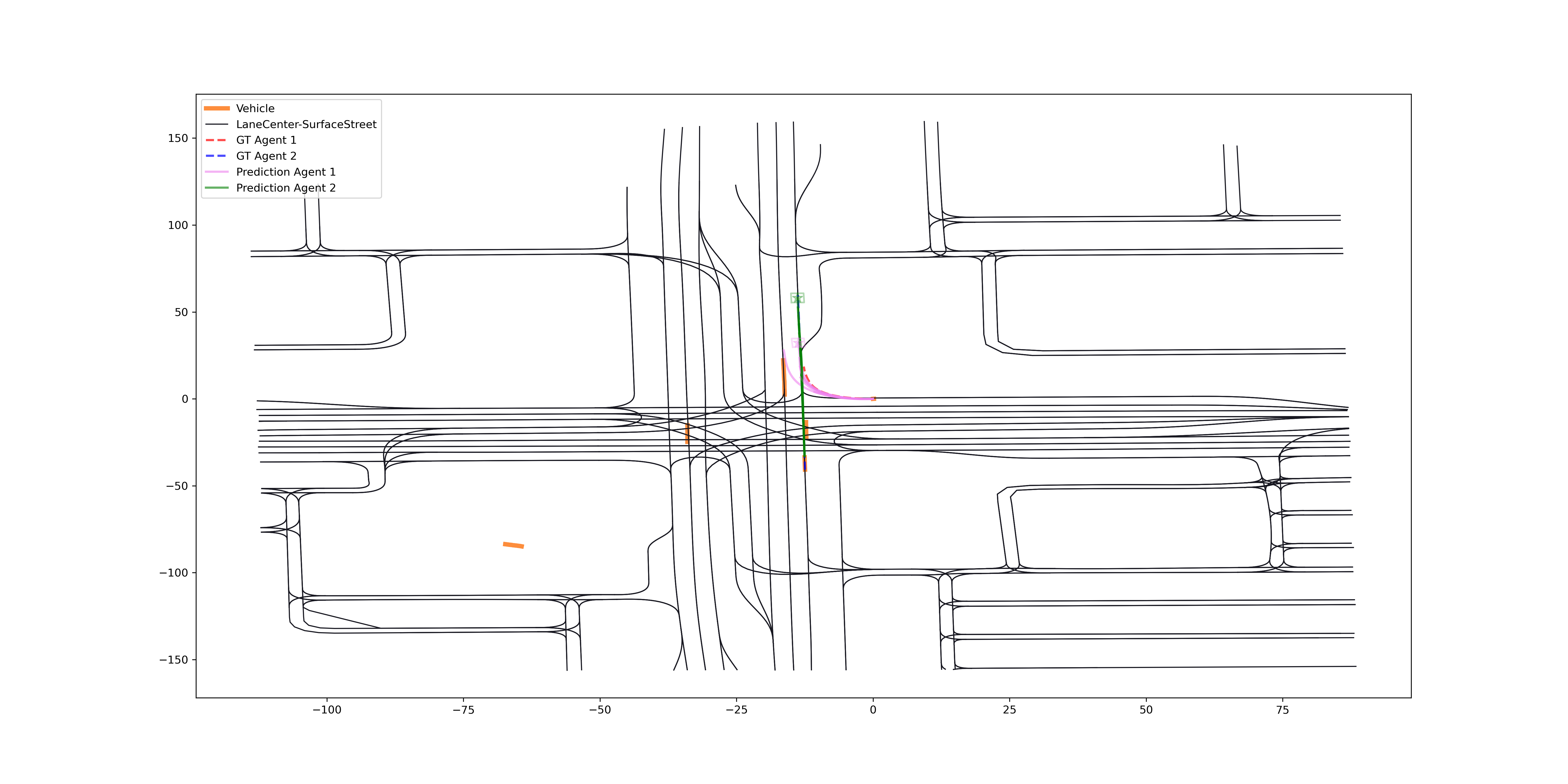}
      \caption{\small Marginal Model Prediction for Scenario: c23ca2936c737633}
    \end{subfigure}
    \hfill
    \begin{subfigure}[b]{0.475\textwidth}
      \includegraphics[width=1\linewidth]{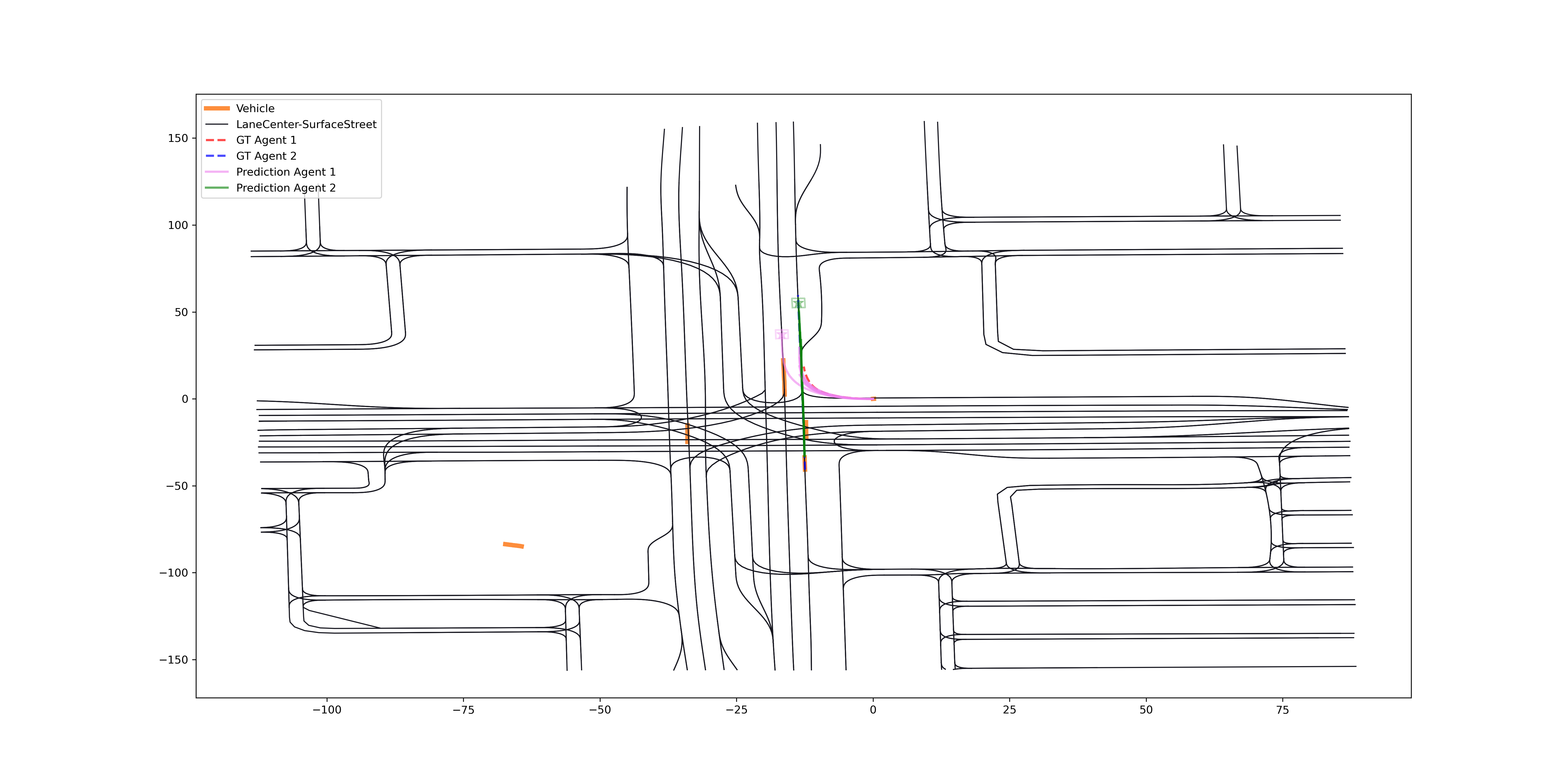}
      \caption{\small Joint Model Prediction for Scenario ID: c23ca2936c737633}
    \end{subfigure}
    \caption{\small The joint model predicts lane change in a two-right-turns scenario while the ground truth is not switching lanes. }
\end{figure*}

\textbf{Joint modal deviates from ground truth due to high interaction.}

\begin{figure*}[ht!]
    \centering
    \begin{subfigure}[b]{0.475\textwidth}
      \includegraphics[width=1\linewidth]{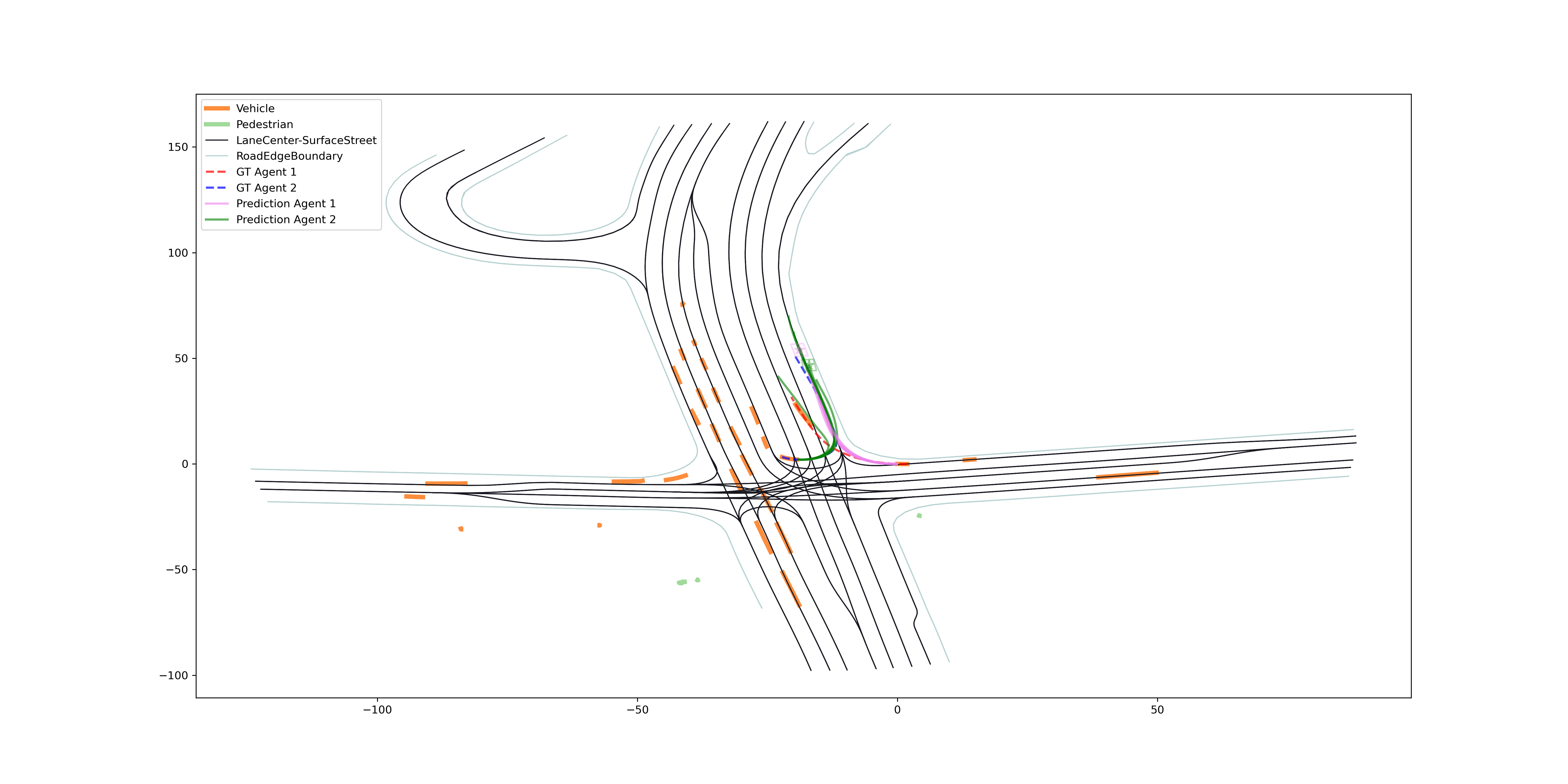}
      \caption{\small Marginal Model Prediction for Scenario: 48aa99ecbb9bbbb3}
    \end{subfigure}
    \hfill
    \begin{subfigure}[b]{0.475\textwidth}
      \includegraphics[width=1\linewidth]{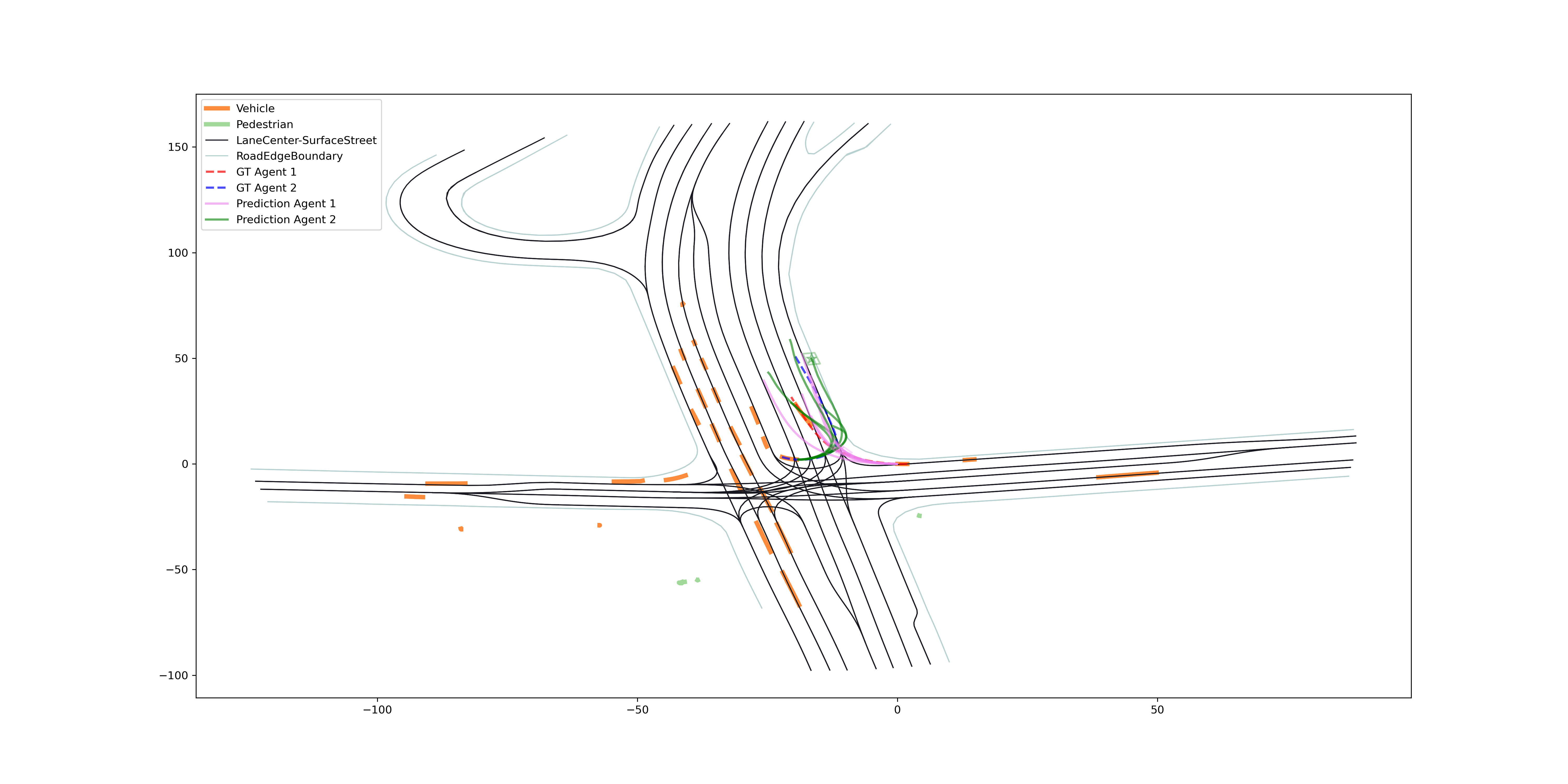}
      \caption{\small Joint Model Prediction for Scenario ID: 48aa99ecbb9bbbb3}
    \end{subfigure}
    \caption{\small The joint prediction is aware of the complicated situation but failed to make a correct prediction. agent A's top mode hesitate before the trajectory intersection point.}
\end{figure*}

\begin{figure*}[ht!]
    \centering
    \begin{subfigure}[b]{0.475\textwidth}
      \includegraphics[width=1\linewidth]{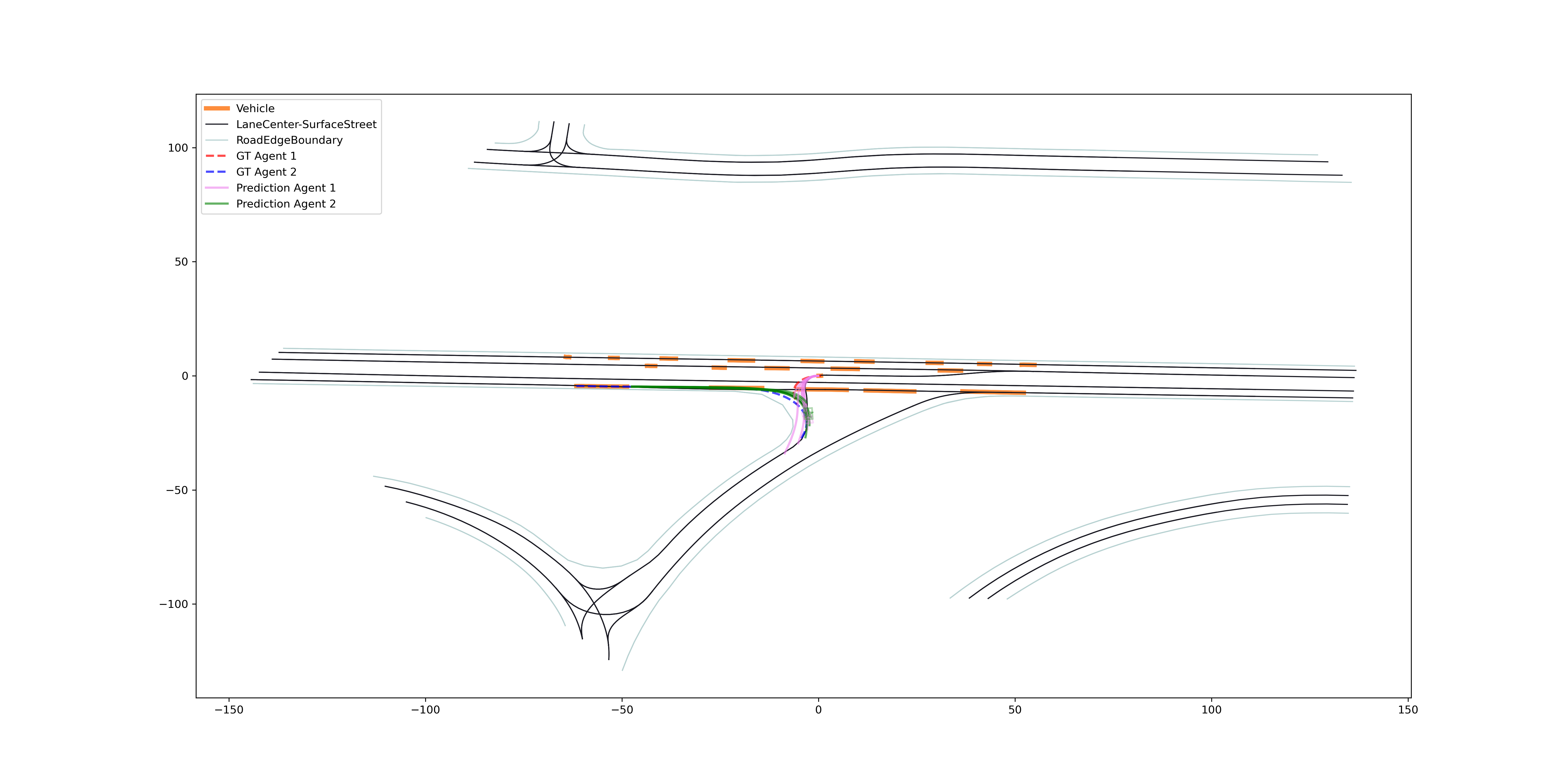}
      \caption{\small Marginal Model Prediction for Scenario: 42cc950c45c6c043}
    \end{subfigure}
    \hfill
    \begin{subfigure}[b]{0.475\textwidth}
      \includegraphics[width=1\linewidth]{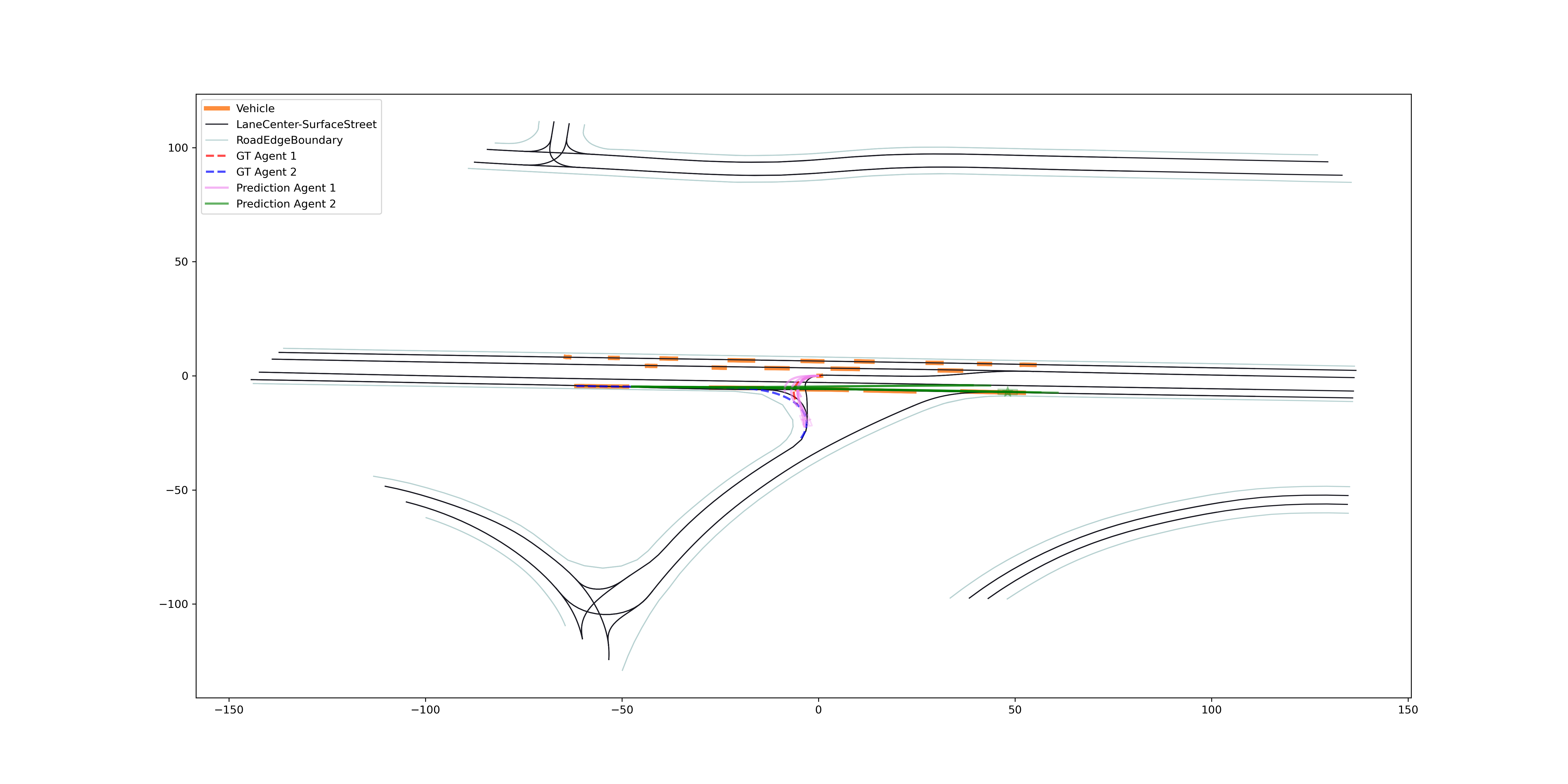}
      \caption{\small Joint Model Prediction for Scenario ID: 42cc950c45c6c043}
    \end{subfigure}
    \caption{\small The joint model forces the prediction to change intent to avoid conflicts but deviates from the ground truth.}
\end{figure*}

\vfill
\end{document}